\documentclass[conference]{IEEEtran}
\IEEEoverridecommandlockouts
\usepackage{cite}
\usepackage{amsmath,amssymb,amsfonts}
\usepackage{algorithmic}
\usepackage{graphicx}
\usepackage{textcomp}
\usepackage{xcolor}
\usepackage{hyperref}
\usepackage{amssymb} 
\usepackage{caption} 
\usepackage{booktabs}
\usepackage{subfigure}
\usepackage{multicol,lipsum}
\usepackage{multirow}
\usepackage{algorithm}
\usepackage{romannum}
\usepackage{balance}

\usepackage{amsmath,amsfonts,bm}

\def\eqref#1{equation~\ref{#1}}

\def\1{\bm{1}}

\def\vn{{\bm{n}}}

\def\vp{{\bm{p}}}

\def\vs{{\bm{s}}}
\def\vt{{\bm{t}}}

\def\vx{{\bm{x}}}
\def\vy{{\bm{y}}}

\def\mM{{\bm{M}}}

\DeclareMathAlphabet{\mathsfit}{\encodingdefault}{\sfdefault}{m}{sl}
\SetMathAlphabet{\mathsfit}{bold}{\encodingdefault}{\sfdefault}{bx}{n}

\def\sD{{\mathbb{D}}}

\def\sN{{\mathbb{N}}}

\def\sP{{\mathbb{P}}}

\def\sS{{\mathbb{S}}}

\newcommand{\R}{\mathbb{R}}

\usepackage[misc]{ifsym}

\newtheorem{myDef}{Definition}

\def\BibTeX{{\rm B\kern-.05em{\sc i\kern-.025em b}\kern-.08em
    T\kern-.1667em\lower.7ex\hbox{E}\kern-.125emX}}

\begin{document}

\title{Weakly-supervised Temporal Path Representation Learning with Contrastive Curriculum Learning --- Extended Version}

\author{Sean Bin Yang$^{1}$, Chenjuan Guo$^{1}$, Jilin Hu$^{1,\text{\Letter}}$
\thanks{Corresponding Author: Jilin Hu}, Bin Yang$^{1}$, Jian Tang$^{2,3,4}$, and Christian S. Jensen$^{1}$ \\
$^1$Department of Computer Science, Aalborg University, Denmark \\
$^2$Mila-Quebec AI Institute $^3$HEC Montreal, Canada $^4$CIFAR AI Research Chair \\

\texttt{\{seany, cguo, hujilin, byang, csj \}@cs.aau.dk}, \texttt{jian.tang@hec.ca} \\}

\maketitle
\thispagestyle{plain}
\pagestyle{plain}
\pagenumbering{arabic}

\begin{abstract}
In step with the digitalization of transportation, we are witnessing a growing range of path-based smart-city applications, e.g., travel-time estimation and travel path ranking. 
A temporal path(TP) that includes temporal information, e.g., departure time, into the path is of fundamental to enable such applications. 
In this setting, it is essential to %
learn generic temporal path representations(TPRs) that consider spatial and temporal correlations simultaneously and that can be used in different  applications, i.e., downstream tasks. 
Existing methods fail to achieve the goal since %
(i) supervised methods require large amounts of task-specific labels when training and thus fail to generalize the obtained TPRs to other tasks; 
(ii) though unsupervised methods can learn generic representations, they 
disregard the temporal aspect, leading to sub-optimal results. 

To contend with the limitations of existing solutions, we propose a Weakly-Supervised Contrastive learning model.
We first propose a temporal path encoder that encodes both the spatial and temporal information of a temporal path into a TPR. 
To train the encoder, we introduce weak labels that are easy and inexpensive to obtain, and are relevant to different tasks, e.g., temporal labels indicating peak vs. off-peak hour from departure times. Based on the weak labels, we construct meaningful positive and negative temporal path samples by considering both spatial and temporal information, which facilities training the encoder using contrastive learning by pulling closer the positive samples' representations while pushing away the negative samples' representations.  
To better guide the contrastive learning, we propose a learning strategy based on Curriculum Learning such that the learning performs from easy to hard training instances. Experimental studies involving three downstream tasks, i.e., travel time estimation, path ranking, and path recommendation, on three road networks offer strong evidence that the proposal is superior to state-of-the-art unsupervised and supervised methods and that it can be used as a pre-training approach to enhance supervised TPR learning. 

This is an extended version of "Weakly-supervised Temporal Path Representation Learning with Contrastive Curriculum Learning"~\cite{SeanIcde2022}, to appear in IEEE ICDE 2022.
\end{abstract}


\section{Introduction}
\label{sec:intro}
\begin{figure}[t]
\centering    
\subfigure[8:00 a.m.] {
 \label{fig:intro_a}     
\includegraphics[height=0.55\columnwidth, width=0.45\columnwidth]{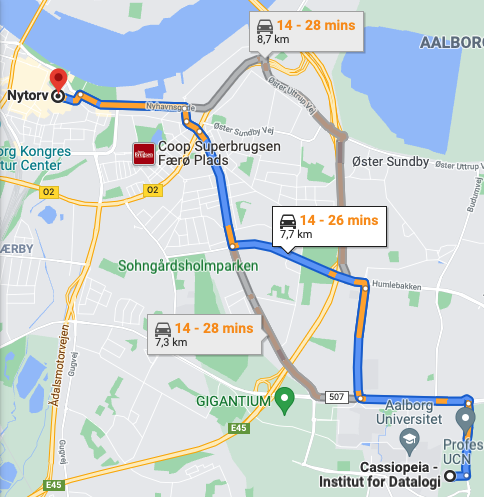} 
} 
\subfigure[10:00 a.m.] { 
\label{fig:intro_b}     
\includegraphics[height=0.55\columnwidth, width=0.45\columnwidth]{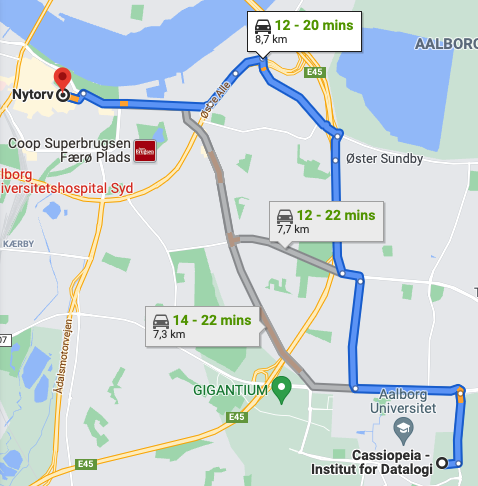} 
}  
\caption{Travel Time Estimation. Travel times of paths from \emph{Cassiopeia} to \emph{Nytorv} at (a) 8:00 a.m. and~(b) 10:00 a.m.
 }
\label{fig:ttds}  
\vspace{-20pt}
\end{figure}

Road-network paths are central in many intelligent transportation system applications, such as path recommendation~\cite{DBLP:journals/vldb/GuoYHJC20,hu2018risk,DBLP:conf/icde/Guo0HJ18}, routing~\cite{DBLP:journals/pvldb/PedersenYJ20, DBLP:conf/icde/LiuJYZ18,DBLP:journals/vldb/PedersenYJ20,zheng2021soup}, travel cost estimation~\cite{DBLP:journals/vldb/YangDGJH18, DBLP:conf/icde/Hu0GJX20,DBLP:conf/cikm/Kieu0GJ18,hu2019stochastic}, and traffic  analysis~\cite{DBLP:conf/icde/CirsteaKG0P21, Razvanicde2022, wupvldb,DBLP:conf/waim/YuanSWYZY10,DBLP:journals/pvldb/CamposKGHZYJ21,tungicde2022,tungicde2022second}.
Path representation(PR) learning is the process of learning representations of paths in the form of vectors with a fixed and relatively low dimensionality that is independent of the actual lengths of path. Thus, such representations can render downstream applications that operate on paths much more efficient than what is possible when operating directly on traditional, variable-length representations of paths. This illustrates the potential of path representation learning for improving intelligent transportation applications. Indeed, initial studies of path representation learning~\cite{DBLP:conf/ijcai/YangGHT021,DBLP:conf/www/LiCSC19} already exist. 

In this study, we aim at learning generic temporal path representations(TPRs), meaning that the representations can be utilized in variety of downstream tasks, and we do so without the need for task-specific labeled data. 
Next the temporal aspect is essential in transportation applications. 
Consider the travel-time estimation example from Google Maps \footnote{\href{http://maps.google.com}{http://maps.google.com}} in Fig.~\ref{fig:ttds}.
Travel from \emph{``Cassiopeia"} to \emph{``Nytorv"} takes longer at 8:00 a.m. than at 10:00 a.m., due to the traffic congestion during morning peak hours. Further, it can be seen that the path recommendation rankings are also different. It recommends to avoid the highway at 8:00 a.m. due to the heavy congestion there, while recommends the highway again at 10:00 a.m., when the traffic is clear. 
Learning path representations without considering the temporal aspect results in poor accuracy, which in turn reduces the utility of such representation in downstream tasks.
However, it is non-trivial to learn generic TPRs using 
either supervised or unsupervised learning.

\begin{figure*}[htp]
\includegraphics[width=\textwidth]{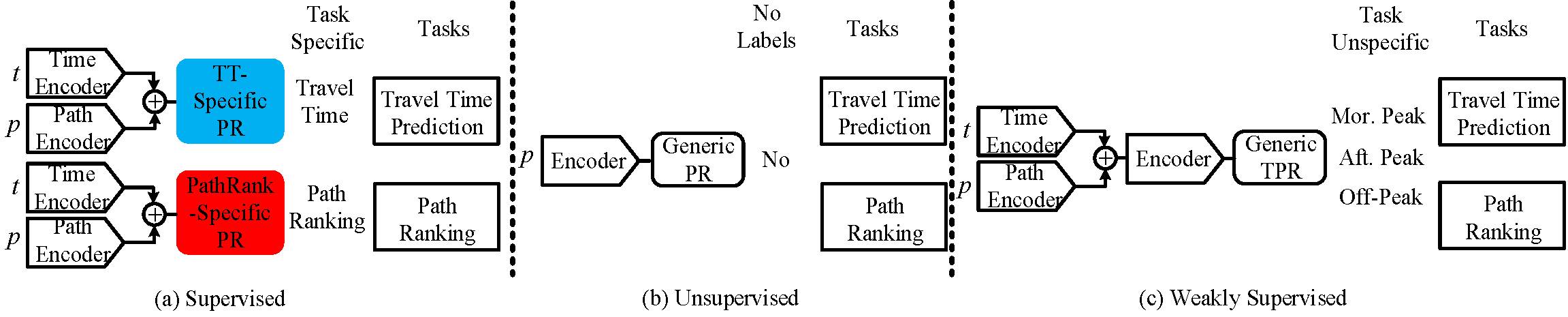}
\caption{Supervised, Unsupervised, and Weakly-Supervised  methods for learning Temporal Path Representations (TPR): (a) Supervised learning relies on task-specific labels to obtain task-specific path representations (PRs), and thus fails to generalize across tasks; (b) Unsupervised learning produces generic path representations for use in different tasks, but fails to capture temporal traffic aspects of paths; (c) Weakly supervised learning (\textbf{Ours}) uses weak labels to learn generic TPRs.}
\label{fig:suw}
\vspace{-15pt}
\end{figure*}

Supervised approaches (Fig.~\ref{fig:suw}(a)) learn TPRs based on task-specific labels~\cite{yang2020context,DBLP:conf/kdd/LiHCSWZP19}. We call these ``strong'' labels because their use targets specific tasks. For example, for the task of travel-time prediction, the time and path encoders first take as input a departure time $t$ and a path $p$, respectively. Then, their outputs are aggregated into a travel-time(TT) specific TPR, which is then utilized to predict the travel time of path $p$ when departure at time $t$.
This supervised learning approach has
two drawbacks:  (\romannum{1}) task specific TPRs do not generalize across tasks. For example, TPRs learned from travel-time labels may perform poorly in path ranking tasks.
(\romannum{2}) supervised learning requires a large amount of labeled training data, which may be impossible or expensive to obtain.

Unsupervised approaches do not rely on task specific labels and are thus able to offer generic path representations. 
Existing unsupervised path representation approaches rely heavily on unsupervised graph representation learning, where the representations of the edges in a path are aggregated into a path representation~\cite{DBLP:conf/ijcai/YangGHT021,DBLP:conf/iclr/SunHV020}. Since existing graph representation learning does not consider temporal information, the obtained path representations also lacks the temporal aspect. 
However, as argued in the context of Fig.~\ref{fig:ttds}, disregarding the temporal aspect adversely affects the quality of downstream tasks. 

In this paper, we target at a solution that is able to offer generic TPRs that take into account spatial and temporal correlations simultaneously without using task specific labels. 
To this end, we propose a temporal path encoder, consisting of a temporal and a spatial embedding module, to encode a temporal path into a TPR by considering both spatial and temporal information. Specifically, we construct a temporal graph to learn temporal embeddings for different departure times via graph representation learning; next, we embed various traffic related information from pertinent road networks into spatial embeddings. 
Finally, the temporal path encoder combines 
the temporal and spatial embeddings to generate the TPR of the input temporal path. 

To enable the training of the temporal path encoder 
such that the obtained TPRs are generic and include temporal information, we introduce weak labels on the temporal aspect. Such weak labels 
are easy to obtain and are relevant to different tasks. Example weak labels include labels indicating peak vs. off-peak periods, which only depend on departure times. 
This way, all temporal paths are associated with weak labels according to their departure times. 

Next, we construct meaningful positive and negative temporal path samples to enable contrastive learning such that no task-specific labels are required and thus the learned TPRs are generic across downstream tasks. The positive samples are those with the same paths and same weak labels
~and all other temporal paths, i.e., same paths with different weak labels, different paths with both same and different weak labels, are negative samples. To learn meaningful representations, we design an objective function to try to pull together representations of positive samples, while separating representations of negative samples. This enables generic TPRs while capturing the temporal information. 
Unlike the supervised methods, we therefore do not require strong, task-specific labeled data for our training. Instead, by deriving weak labels for temporal paths we obtain more generic representations. And unlike the unsupervised approaches, we consider the temporal aspect when we learn path representations. 

To further enhance the weakly-supervised contrastive learning, we integrate curriculum learning strategies to improve the convergence rate and generalization capabilities of the TPR learning. 
Specifically, we propose an curriculum sample evaluation model that outputs difficulty scores for all training samples, according to which the training samples can be sorted. 
To achieve this we first split the training data set into non-overlapping meta-sets. Then, we train separate weakly supervised contrastive(WSC) models on each meta-set, respectively. 
Next, we calculate a TPR similarity score and treat it as a difficulty score for each training sample, based on which we sort all the samples. 
Finally, we provide a curriculum selection algorithm to perform the curriculum learning according to the difficulty scores. 

To the best of our knowledge, this is the first solution that combines advantages of supervised and unsupervised learning to learn generic temporal path representations. In summary, we make the following contributions.
\begin{itemize}
    \item We formulate the temporal path representation learning problem.
    \item We propose a weakly-supervised, contrastive model (basic framework) to learn generic path representations that take temporal information into account.
    \item We integrate curriculum learning into the weakly-supervised contrastive model to further enhance the learned temporal path representations, yields the advanced framework.
    \item We report on extensive experiments using three real-world data sets in the settings of three downstream tasks to assess in detail the effectiveness of the proposed framework.
\end{itemize}

A preliminary version~\cite{SeanIcde2022} has been accepted by the
38th IEEE International Conference on Data Engineering (ICDE 2022). This version offers more detailed technical specifics and additional experimental results. 

\section{Related Work}
\label{sec:rw}
\subsection{Path Representation Learning} 
Deep learning is already being used for representation learning, and different studies have proposed a variety of methods to learn useful path representations. 
To the best of our knowledge, recurrent neural network(RNN) architectures, including long short-term memory(LSTM)~\cite{DBLP:journals/neco/HochreiterS97} and gated recurrent unit(GRU) networks~\cite{DBLP:journals/corr/ChungGCB14}, have been established firmly as the state-of-the-art for path representation learning. Deepcas~\cite{DBLP:conf/www/LiMGM17} leverages bi-directional GRUs to sequentially
process forward and backward node representations of paths, representing a path by the concatenation of resulting forward and backward hidden vectors. ProxEmbed~\cite{DBLP:conf/aaai/LiuZZZCWY17} uses LSTMs to process node representations and apply max-pooling on outputs across all time steps to generate a path representation. SPAE~\cite{DBLP:conf/kdd/LiHCSWZP19} proposes self-attentive path embedding. Paths of arbitrary length are first embedded into fixed-length vectors that are then fed to LSTMs to generate path representations. PathRank~\cite{yang2020context} propose a supervised path representation learning model that takes departure time as additional context information. 
The above methods all perform end-to-end training and rely on the availability of large amounts of labeled training data. In addition, their path representations are task specific.
Most recently, the unsupervised path representation learning framework \emph{PIM}~\cite{DBLP:conf/ijcai/YangGHT021} learns path representations. However, it does not include temporal information. In contrast, we propose a temporal path representation learning framework based on weakly-supervised contrastive loss that can learn path representations when given different departure times.

\subsection{Contrastive Learning}
Recently, the most effective approaches for learning representations with or without labeled data have been supervised or unsupervised contrastive learning~\cite{DBLP:conf/iclr/HjelmFLGBTB19,DBLP:conf/iclr/VelickovicFHLBH19,DBLP:conf/iclr/SunHV020,DBLP:conf/www/PengHLZRXH20,DBLP:conf/ijcai/YangGHT021,DBLP:conf/nips/KhoslaTWSTIMLK20}, which have shown impressive performance in computer vision and graph learning. As a form of metric learning~\cite{DBLP:conf/nips/BromleyGLSS93}, contrastive approaches achieve representations in a discriminating manner through contrasting positive data pairs against negative data pairs. In early work, Hjelm et al.~\cite{DBLP:conf/iclr/HjelmFLGBTB19} proposed Deep InfoMax(DIM) for learning a generic image representations by maximizing mutual information between local and global features in an unsupervised manner. Inspired by DIM, Velickovic et al.~\cite{DBLP:conf/iclr/VelickovicFHLBH19} proposed a similar approach, called Deep Graph Informax(DGI), that learns graph-node representations in an unsupervised manner. Recently, Sun et al.~\cite{DBLP:conf/iclr/SunHV020} proposed InfoGraph for graph representation learning and evaluated the proposal in both unsupervised and semi-supervised settings. 
Most Recently, Khosla et al. ~\cite{DBLP:conf/nips/KhoslaTWSTIMLK20} extended the self-supervised batch contrastive method to a fully-supervised setting, making it possible to leverage label information effectively. However, no previous studies have explored the direction of weakly supervised contrastive learning.  
\begin{table}[tp]
	\caption{{Notations}}
	\centering
	\label{tb:notation}
	\begin{tabular}{l|l}
		\toprule[2pt]
		Notations & Definition \\ \hline
		$G$, $\mathbb{V}$, and $\mathbb{E}$      & Road network, its Vertex set and Edge set          \\ \hline
		$\mathbb{G^\prime}$, $\mathbb{V^\prime}$,  and $\mathbb{E^\prime}$         &Temporal graph, its Vertex set and  Edge set            \\ \hline
		$e_i$ and $v_i$      & $e_i\in \mathbb{E}$ and  $v_i\in \mathbb{V}$ are the $i$-th edge and node in G.           \\ \hline
		$y_i$       & Weak label          \\ \hline
		$\vp$      & Path            \\ \hline
		$tp$      & Temporal path            \\ \hline
		$\vt^{all}$  & Temporal feature embedding \\ \hline
		$\vs^{all}$ & Spatial feature embedding \\ \hline
		$\sS_{tp_i}$ and $\sN_{tp_i}$ &Positive and negative path sets \\ \hline
		$\mathbb{PN}_{i}$ and $\mathbb{NN}_{i}$ & Positive and negative edge sets \\ \hline
		$\mathbb{\sD}$ and $\widetilde{\sD}_{i}$ & Trainig dataset and its $i$-th  meta-set \\ \hline
		$\widetilde{\mathit{WSC}}_{i}$ & Expert model \\ \toprule[2pt]
	\end{tabular}
\end{table}

\subsection{Curriculum Learning}

Inspired by the human learning principle of starting by learning simple tasks before proceeding to learn increasingly hard tasks, curriculum learning(CL)~\cite{DBLP:conf/icml/BengioLCW09} uses nonuniform sampling of mini-batches according to the order of sample difficulty. 
Due its great potential to improve sample efficiency for different deep learning models, CL has attracted considerable interest and has found application in different research domains, e.g., computer vision~\cite{DBLP:conf/cvpr/HuangWT0SLLH20, kong2021adaptive, DBLP:conf/iccv/WangGYWY19}, and natural language processing(NLP)~\cite{DBLP:conf/acl/ShenF20, DBLP:conf/acl/WangWLZY20, DBLP:conf/acl/XuZMWXZ20}.
%
However, none of these studies apply CL to path representation learning. 
\emph{PIM}~\cite{DBLP:conf/ijcai/YangGHT021} is the closet to our paper, in that it proposes a curriculum negative sampling method to enhance the path representation learning. However, \emph{PIM} focuses on negative sampling generations, but not on training. 
Xu et al.~\cite{DBLP:conf/acl/XuZMWXZ20} propose two-staged curriculum learning for NLP, including difficulty evaluation and curriculum arrangement. Inspired by Xu et al.~\cite{DBLP:conf/acl/XuZMWXZ20}, 
we propose a curriculum learning framework that can evaluate the difficulty levels of data in a training data set automatically, so that models can be trained on increasingly difficult subsets of the training data set.
The new framework features two key novelties. (i) Difficulty score computation: In NLP settings, difficulty scores, e.g., accuracy or F1 score, are computed based on strong labels in a supervised setting. In contrast, our difficulty scores are computed based on representation similarities, which do not rely on strong labels. (ii) How the training data is split into metasets: In NLP settings, training data is often split into metasets at random. In contrast, we split the training data based on the lengths of paths. This facilitates distinguishing the difficulty scores of paths.

\section{Preliminaries}
\label{sec:pre}

We first cover important concepts and then present the problem statement.  To ease understanding, Table~\ref{tb:notation} lists important notations that we use throughout this paper.

\subsection{Definitions}

\begin{myDef}
    \textbf{Road network.} A road network is defined as a directed graph $G = (\mathbb{V}, \mathbb{E})$, where $\mathbb {V}$ is a set of vertices $v_i$ that represent intersections and ${\mathbb {E}} \subset \mathbb{V} \times \mathbb{V}$ is a set of edges $e_i=(v_j,v_k)$ that represent edges. Fig.~\ref{fig:rn} shows an example road network.

\begin{figure}[h]
\setlength{\abovecaptionskip}{0.2cm}
\setlength{\belowdisplayskip}{0.2cm}
\centering
\includegraphics[width=0.45\columnwidth]{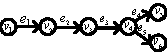}
  \caption{An Example Road Network}  
  \label{fig:rn}
  \vspace{-10pt}
\end{figure}    
\end{myDef}

\begin{myDef}
    \textbf{GPS Trajectory.} A GPS trajectory $\mathit{traj}= \langle (l_i, t_i)
    \rangle_{i=1}^{|\mathit{traj}|}$ of a moving object is defined as a timestamped location sequence, where $l_i$ represents the GPS location at timestamp $t_i$. 
\end{myDef}

\begin{myDef}
     \textbf{Path.} A path $\vp = \langle {e_i} \rangle_{i=1}^{|\vp|}$ is a sequence of adjacent edges, where $e_i \in \mathbb {E}$ is the $i$-th edge in the path. 
\end{myDef}

\begin{myDef}
     \textbf{Temporal Path.} A temporal path is given by ${tp} = (\vp, t)$, where $tp.{\vp}$ is a path and $tp.{t}$ is a departure time.
\end{myDef}

\begin{myDef}
     \textbf{Downstream Task.} A downstream task is a task that  make estimations based on temporal path representations. Specifically, we consider travel time estimation, path ranking score estimation, and path recommendation. 
\end{myDef}

\begin{myDef}
     \textbf{Weak Labels. } Weak labels are easy and inexpensive to obtain, which do not dependent on specific tasks but are relevant to different tasks. 
\end{myDef}

\emph{Example.} An example of weak labels are labels indicating peak v.s. off-peaks periods based on the departure time. For example, it can be Morning peak~(7 to 9 a.m., weekdays), Afternoon peak~(4 to 7 p.m., weekdays), and Off-peak~(all other times). Such labels are easy and inexpensive to obtain, compared to task-specific labels, e.g., labels indicating travel time or path ranking for different paths. Meanwhile, such weak labels are also relevant for three downstream tasks, because the travel time, path ranking, and path recommendation of the same path during peak vs. off-peak periods often differ significantly. 

\begin{myDef}
     \textbf{Temporal Path Representation. } The temporal path representation $\mathit{TPR}_{tp}$ of a temporal path $tp$ is a vector in $\R^{d_h}$, where $d_h$ is the dimensionality of the vector. 
\end{myDef}

\begin{figure}[t]
\centering
\includegraphics[width=0.9\columnwidth]{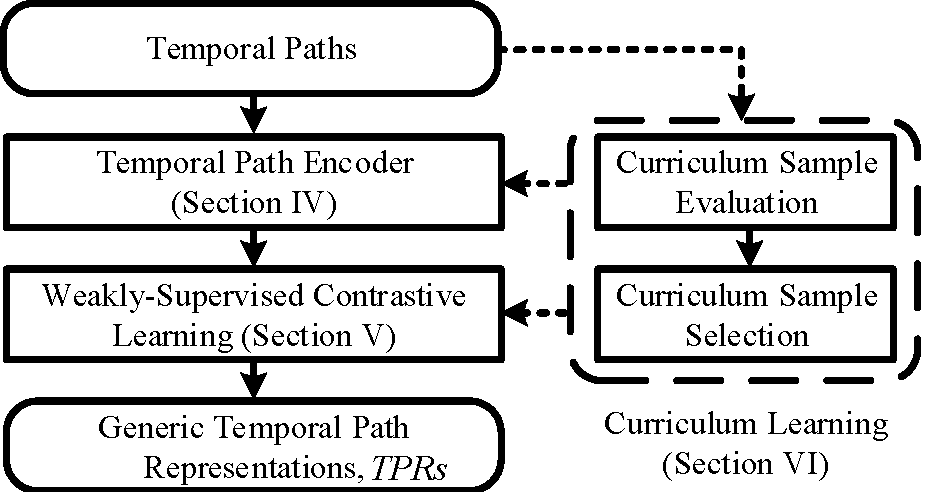}
\caption{Solution Overview}
\label{fig:solver}
\vspace{-15pt}
\end{figure}

\begin{figure*}[t]
\centering
\includegraphics[width=\textwidth]{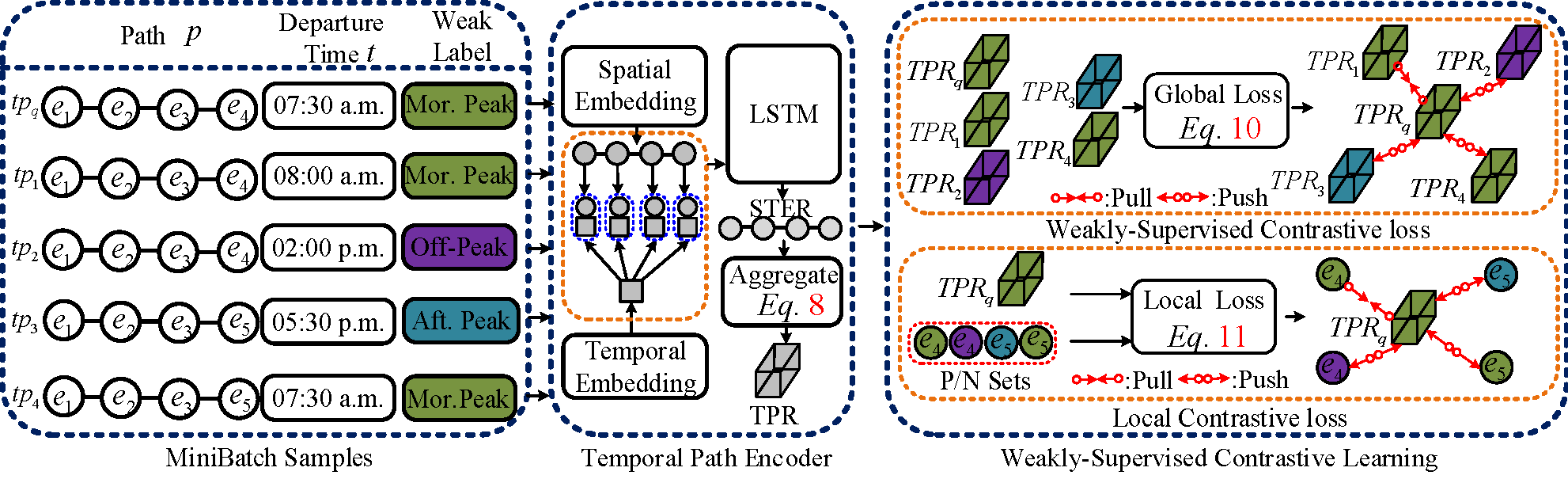}
\caption{Illustration of Basic Framework. Given a set of temporal paths in a minibatch, an input temporal path is encoded into a feature map by the temporal path encoder. The global loss framework takes a (query temporal path $tp_{q}$, positive or negative temporal path) pair as input and pulls together the TPRs of the query path and positive path, while pushing away the TPRs of the query path and negative paths. The local loss takes as input a TPR of query path and a spatio-temporal edge representation (STER) and brings a TPR of a query path with the positive edge representations closer while pushing apart TPR of a query path with the negative edge representations. To illustrate, consider the example with 1 query path $tp_{q}$, 1 positive temporal path $tp_{1}$, and 3 negative temporal paths $tp_{2}$, $tp_{3}$, and $tp_{4}$. Global loss try to pull $\mathit{TPR}_{q}$ with $\mathit{TPR}_{1}$ while push $\mathit{TPR}_{q}$ with $\mathit{TPR}_{2}$, $\mathit{TPR}_{3}$ and $\mathit{TPR}_{4}$. Next we randomly select edges from the positive and negative paths and construct positive/negative edge sets (P/N Sets, i.e., $(e_4, \text{Mor. Peak})$, $(e_4, \text{Off-Peak}$, $(e_5, \text{Aft. Peak})$, and $(e_5, \text{Mor. Peak})$ ). Then, local loss tries to pull $\mathit{TPR}_{q}$ with \textit{STER}$(e_{4}, \text{Mor. Peak})$ and to push $\mathit{TPR}_{q}$ with \textit{STER}$(e_4), \text{Off-Peak})$, \textit{STER}$(e_5), \text{Aft. Peak})$, \textit{STER}($e_5$), \text{Mor. Peak}).}
\label{fig:prs}
\vspace{-15pt}
\end{figure*}

\subsection{Problem Statement}
Given a set of temporal paths $\mathbb{TP}=\{ tp_{1},tp_{2},\ldots, tp_{n} \}$ where each temporal path $tp_i$ is with a weak label $y_i$, 
temporal path representation learning~(\emph{TPRL}) aims at learning a temporal path representation 
$\mathit{TPRL}(tp_i)$ for each temporal path ${tp}_i\in
\mathbb{TP}$ as formulated in Eq.~\ref{eq:pf}. 

\begin{equation}
\label{eq:pf}
    \mathit{TPRL}_{\psi}({tp}_i): \R^{d_{tem}} \times {\mathbb{R}}^{M\times d}  \rightarrow {\mathbb{R}}^{d_h}, 
\end{equation}

\noindent
where $\psi$ represents the learnable parameters for the path encoder, $M$ is the total number of edges in the path, 
$d$, $d_{tem}$, and $d_h$ are the feature dimensions for an edge, %
a departure time embedding, %
and a resulted temporal path representation, respectively. %
\subsection{Solution Overview}
Fig.~\ref{fig:solver} shows an overview of the proposed weakly-supervised contrastive curriculum learning(\emph{WSCCL}), which consists of three modules: 1)~Temporal path encoder, 2)~Weakly-supervised contrastive learning, and 3)~Curriculum Learning. The details of those modules are provided in Sections~\ref{sec:tpe},~\ref{sec:wsc}, and~\ref{sec:adv}, respectively. 

\section{Temporal Path Encoder}
\label{sec:tpe}

Fig.~\ref{fig:prs} gives an overview of the WSC base framework. We detail the Temporal Path Encoder, which consists of a Spatial Embedding layer, a Temporal Embedding layer, and an LSTM layer. Spatial embedding takes as input a sequence of edges and outputs a sequence of spatial feature representations. Temporal embedding takes temporal information as input and converts input to a temporal feature vector. Next, we concatenate the spatial and temporal vector representations and feed the resulting representation to the LSTM model that extracts coupled spatio-temporal relationships and outputs spatio-temporal edge representations. Finally, we aggregate these edge representations to obtain the desired temporal path representations. 

\subsection{Temporal Embedding}
Motivated by Yuan et al.~\cite{DBLP:conf/sigmod/Yuan0BF20}, we construct a temporal graph $G^\prime= (\mathbb{V^\prime}, \mathbb{E^\prime})$, where each node $v'\in \mathbb{V^\prime}$ denotes a departure time slot and each edge $e'\in \mathbb{E^\prime}$ denotes a connection between two time slots. We first split the 24 hours of a day into 5-minute time slots to get 288 time slots. Then, to capture periodicities, we consider the 7 days of a week separately to get a total of 2016 nodes in the temporal graph, where each node represents a time slot and a day of the week. 
Next, we use two one-hot vectors, $\vt_s\in \R^{288}$ and $\vt_w\in \R^{7}$, to denote the initial representations for time slots and days of the week, respectively. For example, the departure time 00:06 a.m. on Monday is represented as $\vt_s=[0, 1, 0, \cdots, 0]$ and $\vt_w=[1, 0, 0, 0, 0, 0, 0]$. 

Therefore, the node representation $\vt_g^{emb}$ of temporal graph $G^\prime$ can be formulated as the concatenation of the two representations: $\vt_g^{emb} = [\vt_s, \vt_w] \in \R^{288 + 7}$. 

To consider the local similarities and weekly periodicities, we draw connections between different nodes in the temporal graph. More specifically, we connect adjacent time slots, indicating that neighboring time ranges should be similar. Further, we connect the adjacent nodes during neighboring days, indicating that time ranges during neighboring days should be similar. Finally, we also connect the time slots between Sunday and Monday. 

Next, we apply a graph representation model, specifically,~\textit{node2vec}~\cite{DBLP:conf/kdd/GroverL16}, to the temporal graph to further learn node representations in the graph. This is expressed in Eq.~\ref{eq:node2vec}.
\begin{equation}
\label{eq:node2vec}
    \vt^{all} = \text{Node2Vec}^{tg}(\vt_g^{emb}), 
\end{equation}
where $\vt^{all} \in \R^{d_{tem}}$ is the finalized temporal representation. 

\subsection{Spatial Embedding}
Recall that a path consists of a sequence of edges, each of which as a number of spatial  features, including, e.g., road types, number of lanes.

\noindent
\paragraph{Capture of Spatial Features} The intuitive way to learn TPRs is to encode the spatial features of all edges in a temporal path into TPRs. We consider the following four types of spatial edge features:  \emph{Road Type(RT):} a categorical value that includes primary, secondary, residential, etc. %
\emph{Number of Lanes(NoL):} a real value that represents the number of traffic lanes in the edge. 
\emph{One Way(OW):} a Boolean that indicates whether the edge is one way or not. %
\emph{Traffic Signals(TS):} a Boolean that indicates whether the edge has one or more traffic signals on the edge or not.%

Next, we represent these different categorical features as one-hot vectors, which can be formulated as $\vs_{RT}^{one}\in \R^{n_{rt}}$, $\vs_{NoL}^{one}\in \R^{n_{l}}$, $\vs_{OW}^{one}\in \R^{n_{o}}$, $\vs_{TS}^{one}\in \R^{n_{ts}}$, where $n_{rt}$, $n_{l}$, $n_{o}$, $n_{ts}$ represent the number of possible values in the four types of features.

Afterward, we leverage embedding matrices to convert these sparse one-hot vectors into dense vectors, which is formulated in Eq.~\ref{eq:embed}.
\begin{align}
\label{eq:embed}
    \vs_{RT}^{emb} = \mM_{RT}^{T} \times \vs_{RT}^{one}&~\text{,}~\vs_{NoL}^{emb} = \mM_{NoL}^{T} \times \vs_{NoL}^{one}, \nonumber \\ 
    \vs_{OW}^{emb} = \mM_{OW}^{T} \times \vs_{OW}^{one}&~\text{,}~\vs_{TS}^{emb} = \mM_{TS}^{T} \times \vs_{TS}^{one},
\end{align}
where $\mM_{RT}\in \R^{n_{rt} \times d_{rt}}$, $\mM_{NoL}\in \R^{n_{l} \times d_{l}}$, $\mM_{OW}\in \R^{n_{o} \times d_{o}}$, and $\mM_{TS}\in \R^{n_{ts} \times d_{ts}}$, where $d_{rt}$, $d_{l}$, $d_{o}$, and $d_{ts}$ are feature dimensions for dense vectors of RT, NoL, OW, and TS, respectively. 

%

Finally, we concatenate all four dense features as final spatial feature embeddings for edge $s_i$, which can be formulated as follows. 
\begin{equation}
    \vs^{type}=[ \vs_{RT}^{emb}, \vs_{NoL}^{emb}, \vs_{OW}^{emb}, \vs_{TS}^{emb}],
\end{equation}

\noindent
where $[\cdot, \cdot]$ denotes concatenation vectors. 

\noindent
\paragraph{Road Network Topology} Since each edge has effects on its neighboring edges, connected edges should have similar representations. 
Inspired by graph embedding~\cite{DBLP:conf/kdd/GroverL16, DBLP:conf/www/PengHLZRXH20, DBLP:conf/iclr/VelickovicFHLBH19}, 
which aims to learn node representations in a graph by considering the graph topology. 
Again, we apply \textit{node2vec}~\cite{DBLP:conf/kdd/GroverL16} to learn graph representation of road network, which can be formulated as $\vn_{vi}^{rn} = \text{Node2Vec}^{rn}(\vn_{vi}^{one})$, 
where $\vn_{vi}^{one}$ is one-hot vector of node $v_i$ and $\vn_{vi}^{rn} \in \R^{\frac{d_{top}}{2}}$ is the finalized node representation in a road network. 
The finalized edge representation of $e_k \in \mathbb{E}$ can be formulated as follows. 
\begin{equation}
    \vs_{e_k}^{rn}=[\vn_{v_i}^{rn}, \vn_{v_j}^{rn}],
\end{equation}
where $v_i$ and $v_j$ are start and end nodes of edge $e_k$. 

Finally, the finalized topology feature with spatial feature of edge $e_k$ can be rewritten as follows. 
\begin{equation}
    \vs_{e_k}^{all}=[ \vs_{e_k}^{rn}, \vs^{type}],
\end{equation}
where $\vs_{e_k}^{all}\in \R^{d}$ is the finalized spatial embedding for $e_k$,  where $d=d_{rt}+d_l+d_o+d_{ts}+d_{top}$.
\noindent

\subsection{LSTM Encoder}
Given an input temporal path $tp$, we achieve a sequence of spatio-temporal representations $\langle \vx_{e_{1}},\vx_{e_{2}},\cdots \vx_{e_{\vp}} \rangle$, where $\vx_{e_i}=[ \vt^{all}, \vs_{e_i}^{all}]$, where $\vt^{all}$ is the temporal embedding for $tp.t$. As recurrent neural networks(RNN) are known to be effective at modelling sequences, we feed this spatio-temporal representation into an RNN to further learn path representations. Specifically, we employ an LSTM model~\cite{hochreiter1997long} to capture the sequential dependencies by taking each element of the spatio-temporal representations as input, which can be formulated as follows. 
\begin{equation}
    \mathbf{\hat{p}} = \langle \vec{p}_{e_{1}},\vec{p}_{e_{2}},\cdots \vec{p}_{e_{\vp}} \rangle = \text{LSTM}(\langle \vx_{e_{1}},\vx_{e_{2}},\cdots \vx_{e_{\vp}} \rangle),
\end{equation}
where $\vec{p}_{e_{j}}\in \R^{d_h}$ is the finalized spatio-temporal representation of edge $e_{j}$. $\text{LSTM}(\cdot)$ is the RNN model to capture the sequential dependencies, but it is also possible to use more advanced sequential models, e.g., Transformer~\cite{DBLP:conf/nips/VaswaniSPUJGKP17}.

\subsection{Aggregate Function} 
As shown in Fig.~\ref{fig:prs}, the~\emph{Aggregate} function takes as input a sequence of spatio-temporal edge feature vectors and returns a TPR. In particular, we aggregate $ \mathbf{\hat{p}}$ into a TPR via an aggregate function $\mathcal{P}(\cdot): \mathbb{R}^{n \times d_{h}} \longmapsto \mathbb{R}^{d_{h}}$, where $n$ is the number of edges in the path. We use an average aggregate function that takes the average of the edge representations in $\mathbf{\hat{p}}$ across edges.
\begin{equation}
    \vec{h}_{p}=\frac{\sum_{i=1}^{n}\vec{p}_{i}}{|\mathbf{\hat{p}|}}  \in \mathbb{R}^{d_{h}},
\end{equation}
\noindent
where $\vec{h}_{p}$ represents the temporal path representation, $\vec{p}_{i}$ is the latent representation of edge $i$ in the path. 
%

\section{Weakly-supervised Contrastive Learning}
\label{sec:wsc}
To ensure that we obtain generic TPRs that apply to different downstream tasks, we employ contrastive learning to construct the learning objectives for the whole framework. 
Here, we first detail positive and negative sample generation with weak labels. Then, we show how to construct weakly-supervised contrastive global and local losses. 

\noindent
\subsection{Generation of Positive and Negative Samples} 
\label{sec:pnsg}
Positive and negative samples play an essential role in contrastive learning. 
Contrastive learning does not require strong labels and provides us with a good way of constructing the learning objectives for our model. 
In self-supervised contrastive learning, positive samples are always derived from the same object with different views, e.g., cropped parts of an object or generated by different models. Negative samples are simply representations that come from different objects. 
However, if some negative samples that have properties that are similar to those of positive samples, self-supervised contrastive learning faces difficulties because all negative samples are treated equally. 

Suppose we have a set of temporal paths, positive TPs are not only different representations of the same temporal path, but they also include TPs that traverse the same path with the same weak label. 
In contrast, negative TPs belong to three categories: (\romannum{1}) same paths but different weak labels; (\romannum{2}) different paths but the same weak labels; (\romannum{3}) different paths and different weak labels. 
Therefore, we can generate multiple positive and negative TPs for a query TP. 

The block of MiniBatch Samples in Fig.~\ref{fig:prs} shows an example, with five TPs, i.e., $tp_q, tp_1,tp_2,tp_3,\text{and }tp_4$, and three weak labels, i.e, morning peak (Mor. Peak), Afternoon peak (Aft. Peak) and Off-Peak. If we take $tp_q$ as the query TP, $tp_1$ is the corresponding positive sample since the two share the same path~(i.e., $\langle e_1, e_2, e_3, e_4 \rangle$) and the same departure weak label~(i.e., Mor. Peak), although their exact departure times are different. Next, $tp_2, tp_3, \text{and }tp_4$ are negative samples, where $tp_2$ has the same path but a different weak label, $tp_3$ has a different path and a different weak label, and $tp_4$ has a different path but the same weak label. 

\noindent
\subsection{Global Weakly-supervised Contrastive Loss} 

Given a batch of training samples with batch size $B$, self-supervised contrastive loss can be formulated as in Eq.~\ref{eq:cl}.
\begin{align}
\mathcal{L}^{\text {self }}=\sum_{i=1}^{2 B}-\log \frac{\exp \left(<\Psi\left(x_{i}^{\prime}\right), \Psi\left(x_{p}^{\prime}\right)> / \Psi\left(x_{k}^{\prime}\right)> / \hat{\tau}\right)}{\sum_{k=1}^{2 B} \mathbf{1}_{i \neq k} \exp <\left(\Psi\left(x_{i}^{\prime}\right) , \Psi\left(x_{k}^{\prime}\right)> / \hat{\tau}\right)},
\label{eq:cl}
\end{align}
\noindent
where $\text{exp}(\cdot)$ denotes the exponential operation, $<\cdot,\cdot>$ denotes the inner product of two vectors, $\Psi(\cdot) \in \mathbf{R}^{d_h}$ represents the output from the encoder, $\hat{\tau}$ is a temperature parameter, $x_{p}^{\prime}$ is an alternative view of object $x_{i}^{\prime}$ which can be generated by using data augmentation, e.g., rotate an image by 90 degree. $x_{k}^{\prime}$ denotes the negative samples from the batch, which is all samples from the batch other than $x_{i}^{\prime}$, meanwhile  $\mathbf{1}_{i \neq k}$ is an indicator vector, where all elements are 1s, except a 0 at the $i$-th position. 

However, the self-supervised contrastive loss in Eq.~\ref{eq:cl} is unable to take into account differences among negative samples. Motivated by~\emph{SupCon}~\cite{DBLP:conf/nips/KhoslaTWSTIMLK20}, good generalization requires the ability to capture the similarity between samples in the same class and contrast them with samples in other classes. In this paper, we propose instead a WSC loss that utilizes positive and negative sample generation, as introduced in Section~\ref{sec:pnsg}. As is shown in Fig.~\ref{fig:prs}, for each query temporal path $tp_q$, we try to pull closer TPRs with positive temporal path samples, which can be represented as $\mathit{TPR}_{q} \rightarrow \leftarrow \mathit{TPR}_{1}$, and push away TPRs from negative temporal path samples, denoted by $\mathit{TPR}_{q} \leftarrow \rightarrow \mathit{TPR}_{2}$, $\mathit{TPR}_{q} \leftarrow \rightarrow \mathit{TPR}_{3}$, and $\mathit{TPR}_{q} \leftarrow \rightarrow \mathit{TPR}_{4}$.
This yields global WSC formulated in Eq.~\ref{eq:wsc_loss}. 
\begin{equation}
\label{eq:wsc_loss}
\small
\begin{split}
    & \mathcal{L}^{global} 
    =\sum_{(\boldsymbol{tp}_i, \vy_i) \in \sP} \mathcal{L}_{(\boldsymbol{tp}_i, \vy_i)}^{global} = \sum_{(\boldsymbol{tp}_i, \vy_i) \in \sP} \frac{1}{|\sS_{\boldsymbol{tp}_i}|} \\ 
    &\sum_{\boldsymbol{tp}_j \in \sS_{\boldsymbol{tp}_i}} \log \frac{\exp \left(\text{sim}(\mathit{TPR}_{i}, \mathit{TPR}_{j}) \right)}{\sum_{\boldsymbol{tp}_k \in \sN_{tp_i}} \exp \left(\text{sim}(\mathit{TPR}_{i}, \mathit{TPR}_{k})\right)}, 
\end{split}
\end{equation}

\noindent
where $\text{sim}(\cdot)$ is cosine similarity function that quantifies the similarity between two TPRs; $\sP=\{(tp_i, y_i)\}_{i=1}^{B}$ is a set of temporal paths in one training batch, where $y_i$ is the departure weak label for $tp_i$; $\sS_{tp_i}=\{tp_j\}_{j=1}^{|\sS_{tp_i}|}$ is the positive sample set for query $tp_i$, where $y_i=y_j$ and $tp_i.\textbf{p} = tp_j.\textbf{p}$; and $\sN_{tp_i}=\sP \setminus \{\boldsymbol{tp}_i \cup \sS_{\boldsymbol{tp}_i} \}$ is the negative sample set for query $tp_i$.  

\subsection{Local WSC Loss}

In addition to the weakly-supervised learning across query temporal path with global positive and negative temporal paths, we also consider local differences between positive and negative temporal edge samples. These acts as a strong regularization that enhances the learning ability of our method. The Local Contrastive Loss element in Fig.~\ref{fig:prs} illustrates the design of our local contrastive loss, which consists of WSC with $\mathit{TPR}_{tp_q}$. 

The goal of local contrastive learning is to preserve the local similarity between a TPR and the spatio-temporal representation of its edges. In particular, it is expected that a TPR can capture local similarity (edge-level similarity), i.e., TPRs are close to embeddings of positive edge samples and are distant from embeddings of negative edge samples. Similar to global WSC, we formulate the local WSC loss as maximizing the similarity with positive temporal edge samples as well as minimizing the similarity with negative temporal edge samples.

We proceed to describe the construction of the positive and negative edge samples. First, we randomly select edges that appear in positive temporal paths as our positive edge set, which is denoted as $\mathbb{PN}_{i}$. Then, edges that appear in negative temporal paths are selected as our negative edge set, denoted as $\mathbb{NN}_{i}$. 
Next, we set the weak temporal label of each temporal path be the label of the corresponding edge in the temporal path. 
As is shown in Fig.~\ref{fig:prs}, for each query path $tp_q$, we try to pull TPRs with positive edges closer, e.g., $(\mathit{TPR}_{q}, \text{Mor. Peak}) \rightarrow \leftarrow (\textit{STER}(e_4),\text{Mor. Peak})$, and push away TPRs from negative edges, e.g., ($\mathit{TPR}_{q}$, \text{Mor. Peak}) $\leftarrow \rightarrow$ (\textit{STER}($e_4$), \text{Off-Peak}), ($\mathit{TPR}_{q}$, \text{Mor. Peak}) $\leftarrow \rightarrow$ (\textit{STER}($e_5$), \text{Aft. Peak}), and ($\mathit{TPR}_{q}$, \text{Mor. Peak}) $\leftarrow \rightarrow$ (\textit{STER}($e_5$), \text{Mor. Peak}).

In this phase, the objective of local contrastive learning is to increase the similarity of TPRs with positive edge samples while decreasing the similarity of TPRs with negative edge samples. Using cosine similarity $s(x, y)=x^{\top} y /\|x\|\|y\|$, we aim to optimize the objective function for local contrastive loss that is formulated in Eq.~\ref{eq:local_loss}.
\begin{equation}
\label{eq:local_loss}
\small
\mathcal{L}^{local}  = \sum_{(\boldsymbol{tp}_i, \vy_i) \in \sP} \frac{1}{|\mathbb{PN}_{i}|}\log \frac{\sum_{(\boldsymbol{e}_j, y_{i)} \in \mathbb{PN}_{i}}\exp \left(s(\mathit{TPR}_{i}, e_{j}) \right)}{\sum_{(\boldsymbol{e}_k, y_j \neq y_i) \in \mathbb{NN}_{i}} \exp \left(s(\mathit{TPR}_{i}, e_{k})\right)}, 
\end{equation}
\noindent
where $\mathbb{PN}_{i}$ and $\mathbb{NN}_i$ are the positive and negative edge sets, and $y_i$ denotes the weak label for edge representation, which inherits from the corresponding temporal path.  

\subsection{Objective for WSC}
To train our temporal path encoder in an end-to-end manner, we jointly leverage both the the weakly-supervised global and local contrastive loss. Specifically, the overall objective function to maxmize is defined in Eq.~\ref{eq:jl}.
\begin{equation}
\label{eq:jl}
    \mathcal{L} = \underset{\psi}{\arg \max } \sum_{i \in I} \lambda \cdot \mathcal{L}_{i}^{global}+ (1-\lambda) \cdot \mathcal{L}^{local}_{i}, 
\end{equation}
\noindent
where $\lambda$ is a balancing factor and $I$ is the set of training batches. 

\section{Contrastive Curriculum Learning}
\label{sec:adv}

When training WSC using randomly shuffled training data, the training process is prone to getting stuck in a bad local optimum, which leads to suboptimal TPRs. To alleviate this problem, we build on the intuition that the algorithm should be presented with the training data in a meaningful order that facilitates learning. Specifically, the order of the samples is determined by how easy they are, as this can be expected to enhance weakly-supervised contrastive learning. We proceed to integrate curriculum learning with WSC, thus obtaining the advanced framework called \emph{WSCCL}. 
\begin{figure*}[htp]
\includegraphics[width=\textwidth]{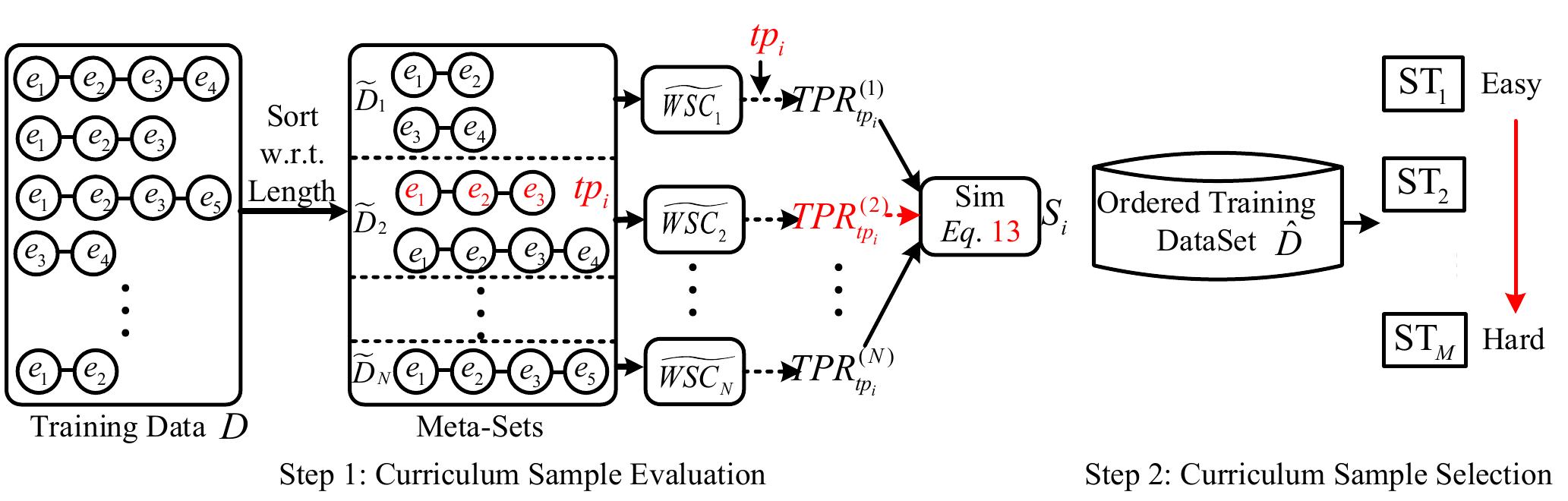}
\caption{Illustration of Advanced Framework}
\label{fig:af}
\vspace{-21pt}
\end{figure*}

\subsection{Overview of Curriculum Learning}

Motivated by Xu et al.~\cite{DBLP:conf/acl/XuZMWXZ20}, we decompose curriculum sample generation into two stages: \emph{Curriculum Sample Evaluation} and \emph{Curriculum Sample Selection}, as shown in Fig.~\ref{fig:af}. 

1) In curriculum sample evaluation, we assign a difficulty score $S_i$ to path $p_{i}$ in the training dataset $\sD$. The score reflects the difficulty of the model to learn a good representation w.r.t. path $p_{i}$. 
2) In curriculum sample selection, we aim to partition the training data into different difficulty stages. 
More specifically, we first sort the training data according to the difficulty scores. Then, we split the sorted training data into a sequence of sorted learning stages $\lbrace \mathit{ST}_i | i = 1, 2, \dots, M\rbrace$ in an easy-to-difficult fashion. Finally, our base model, WSC, is trained according to this curriculum. 
We detail these two stages in Sections~\ref{sec:cse} and~\ref{sec:css}, respectively.

\subsection{Curriculum Sample Evaluation}
\label{sec:cse}
The difficulty of a given temporal path can be quantified in many different ways.
We argue that difficulty scores, like the intrinsic properties of the training dataset, should be decided by the model itself.

We first sort the training data set $\sD$  according to the lengths of the paths. Then, we split $\sD$ into $N$ non-overlapping meta-sets, i.e., $\sD= \lbrace \widetilde{\sD}_{1}, \widetilde{\sD}_{2}, \cdots, \widetilde{\sD}_{N} \rbrace$, where $\widetilde{\sD}_{i}$ is the $i$-th meta-set, $\widetilde{\sD}_{i} \cap \widetilde{\sD}_{j} = \emptyset, \forall i\neq j, i, j\in [1, N]$. 
Next, we train $N$ independent WSC models w.r.t. the different meta-sets, i.e., $\widetilde{\mathit{WSC}}_{i}, i = [1, N]$. 
More specifically, $\widetilde{\mathit{WSC}}_{i}$ is trained only on the $i$-th meta-set, $\widetilde{\sD}_{i}$. 
After that, we obtain a total of $N$ trained independent WSC models, which we call \textit{Experts}, to evaluate the difficulty of each training sample. 

We then use the \textit{Experts} to calculate the difficulty scores for the training data. We take a temporal path $\boldsymbol{tp}_i \in \widetilde{\sD}_{j}$ as input to each \textit{Expert}, $\widetilde{\mathit{WSC}}_{j}, j \in [1,N]$, and obtain a total of $N$ TPRs, i.e., $\mathit{TPR}_{\boldsymbol{tp}_{i}}^{(1)}, \mathit{TPR}_{\boldsymbol{tp}_{i}}^{(2)}, \cdots, \mathit{TPR}_{\boldsymbol{tp}_{i}}^{(N)}$. Since $\boldsymbol{tp}_i$ comes from $\widetilde{\sD}_{j}$, we take $\mathit{TPR}_{\boldsymbol{tp}_{i}}^{(j)}$ as the ground truth, and calculate the similarity between $\mathit{TPR}_{\boldsymbol{tp}_{i}}^{(j)}$ and $\mathit{TPR}_{\boldsymbol{tp}_{i}}^{(k)}$, where $\forall k\neq j, k\in [1, N]$. The resulting similarity scores are then summed up to denote as the difficulty score of temporal path $\boldsymbol{tp}_i$, denoted by $S_{i}$. This is formulated in Eq.~\ref{eq:sim_cl}. 
\begin{align}
\label{eq:sim_cl}
S_{i} =\sum_{k=1,k\neq j}^{N}Sim(\mathit{WSC}_{j}(tp_i), \mathit{WSC}_{k}(tp_i)),
\end{align}
\noindent
where $Sim(\cdot, \cdot)$ denotes the similarity function.

Finally, we repeat the difficulty score calculation for the entire training data. This yields a temporal path dataset $\sD = \{(\boldsymbol{tp}_{1}, S_{1}), \cdots, (\boldsymbol{tp}_{N}, S_{N})\}$, where $(\boldsymbol{tp}_{i}, S_{i})$ is one element and $S_{i}$ is the difficulty score for temporal path $\boldsymbol{tp}_{i}$. 
The intuition is that if one element, i.e., $(\boldsymbol{tp}_{i}, S_{i})$ can obtain a similar representations in all other \textit{Experts} compared to the representation from its own \textit{Expert}, then we denote it as an easy sample. Since we calculate the sum of similarities, the higher $S_{i}$ is, the easier the temporal path $\boldsymbol{tp}_i$. 

\subsection{Curriculum Sample Selection}
\label{sec:css}
To define the curriculum sample selection strategy, we first represent the learning curriculum in a multi-stage manner: $\lbrace \mathit{ST}_{i}| i = 1, 2, \dots, M\rbrace$. More specifically, we rank all the training samples according to their difficulty scores and then distribute them evenly among $M$ training stages. 
This way, the training data is partitioned into $M$ parts with different levels of difficulty, ranging from $S_1$ (the easiest) to $S_M$ (the hardest). To ensure some local variations, the training samples in different stages are shuffled. 

Next, we train our WSC for one epoch at each stage. When the training reaches stage $S_M$, the WSC should be ready for the original distribution in the whole training dataset $\sD$. Finally, we add another stage, denoted by $S_{M+1}$, that covers the entire training data, and WSC is trained on this stage until it converges. For simplicity, we keep $N=M$, as is done elsewhere~\cite{DBLP:conf/acl/XuZMWXZ20}, and we leave the investigation of different combinations of $N$ and $M$ for future work.

\section{Experimental Study}
\label{sec:exp}
\subsection{Experimental Setup}
\subsubsection{Data sets} We use three traffic data sets to study the effectiveness of the proposed framework. Tab.~\ref{tb1:statis} shows the data statistics. 
Using each of these, we report results for three downstream tasks: travel time prediction, path ranking, and path recommendation.

\noindent
\paragraph{Aalborg, Denmark~\cite{yang2020context}} We use the road network graph of Aalborg from OpenStreetMap\footnote{\href{https://www.openstreetmap.org}{https://www.openstreetmap.org}} that consists 10,017 nodes and 11,597 edges. We use a substantial GPS data set that captures travel in this road network. Specifically, the data set consists of 180 million GPS records from 183 vehicles  sampled at 1 Hz over a two-year period. A well-known map matching method~\cite{DBLP:conf/gis/NewsonK09} is used to map match the trajectories. This yields a total of 28,370 paths. 

\noindent
\paragraph{Harbin, China~\cite{DBLP:conf/www/LiCSC19}} The data set was collected from 13,000 taxis in Harbin, China. We extract the corresponding road network from OpenStreetMap. The network contains 8,497 nodes and 14,497 edges. The GPS data was sampled at about 1/30 Hz. 
After map matching, we obtain 58,977 paths.

\noindent
\paragraph{Chengdu, China} The data set was collected in Chengdu, China during October and November 2016\footnote{\href{https://outreach.didichuxing.com/research/opendata/en/}{https://outreach.didichuxing.com/research/opendata/en/}}. We extract the corresponding road network from OpenStreetMap. The network contains 6,632 nodes and 17,038 edges. The GPS data was sampled at about 1/4--1/2~Hz.
After map matching, we obtain 57,404 paths.
\begin{table*}[tp]
	\caption{Statistics of Datasets}
	\centering
	\begin{tabular}{l|lll}
		\toprule[2pt]
		DataSet       & Aalborg, Denmark & Harbin, China & Chengdu, China \\ \toprule[1pt]
		Time Period         &01/012007-30/12/2008               &03/01/2015-07/01/2015     &01/10/2016-031/11/2016               \\ 
		Unlabeled Paths  &28,370               &58,977          &57,404        \\ 
		Labeled Paths   &15,000              &15,000  &15,000\\
		\#Node        &10,017               & 8,497         &6,632        \\ 
		\#Edges       &11,597               & 14,497        &17,038        \\ \toprule[2pt]
	\end{tabular}
	\label{tb1:statis}

\end{table*}

\subsubsection{Downstream Tasks} We consider three tasks.

\noindent
\paragraph{Path Travel Time Estimation} Each path is associated with a travel time (in seconds) obtained from the corresponding trajectory.
We aim at building a regression model to estimate the travel times of paths based on their TPRs. 
We evaluate the accuracy of estimations using the %
Mean Absolute Error (\textbf{MAE}), Mean Absolute Relative Error (\textbf{MARE}), and Mean Absolute Percentage Error (\textbf{MAPE}). Smaller values indicate higher estimation accuracy. These metrics are defined in Eq.~\ref{eq:mae}.
%
\begin{align}
	\label{eq:mae}
	&\operatorname{MAE}(x, \hat{x})=\frac{1}{n} \sum_{i}^{n}\left|x^{(i)}-\hat{x}^{(i)}\right|,\nonumber \\
	& \operatorname{MARE}(\boldsymbol{x}, \dot{\boldsymbol{x}})=\frac{\sum_{i}^{n}\left|x^{(i)}-\hat{x}^{(i)}\right|}{\sum_{i}^{n}\left|x^{(i)}\right|},  \\\nonumber
	& \operatorname{MAPE}(x, \hat{x})=\frac{1}{n} \sum_{i}^{n}\left|\frac{x^{(i)}-\hat{x}^{(i)}}{x^{(i)}}\right|,
\end{align}
where $x_{i}$ and $\hat{x}_{i}$ represent the ground truth travel time and the estimated travel time, respectively; and $n$ is the total number of estimations.

\noindent
\paragraph{Path Ranking}
In path ranking, each path is associated with a ranking score in the range $[0, 1]$. The ranking scores are obtained with the help of historical trajectories by following an existing procedure~\cite{yang2020context}. %
Given a historical trajectory of a driver, we consider the path used by the trajectory, called the trajectory path, as the top ranked path. Then, we use a path finding algorithm to generate multiple paths connecting the same source and destination. We use the similarity between a generated path and the trajectory path for ranking: the more similar a generated path is to the trajectory path, the higher its similarity score; and the trajectory path itself has score 1 and thus ranks the highest.
As for the previous task, we aim at building a regression model to estimate the ranking scores of paths.

To quantify the performance of path ranking, we report the \textbf{MAE} of the estimated ranking scores, the Kendall rank correlation coefficient ($\tau$)~\cite{kendall1938new}, and the Spearman's rank correlation coefficient ($\rho$)~\cite{zar1972significance}.
The latter two capture the similarity, or consistency, between the ground truth and estimated rankings. 
The higher the $\tau$ and $\rho$ values are, the more consistent the two rankings are, indicating higher accuracy.
Specifically, we have the following formulations in Eq.~\ref{eq:pho}. 
\begin{align}
	\label{eq:pho}
	\tau=\frac{{N}_{con} - {N}_{dis}}{n(n-1) / 2}; \\\nonumber
	\rho=1-\frac{6\sum_{i=1}^n d_i^2}{n(n^2-1)}, 
\end{align}
where $N_{con}$ and $N_{dis}$ denote the number of path pairs are consistent and inconsistent in the two rankings. $d_i$ represents the rank difference on the $i$-th competitive path in both rankings.

\noindent
\paragraph{Path Recommendation}
A similar strategy is used in an existing study~\cite{DBLP:journals/pvldb/0026HFZL020}, where a path is associated with a binary label with the help of users’ trajectories. A path used by a user's trajectory, say path A, is labeled 1, whereas alternative paths connecting the same source and destination, say paths B and C, are labeled 0. This follows the intuition that given three paths A, B, and C, the user should choose A, not B and C, meaning that path A should be recommended to the user. We conduct the path recommendation task on all three data sets.
We evaluate the recommendation effectiveness using the classification \textbf{Accuracy} (Acc.) and \textbf{Hit Rate} (HR). Higher values indicate better performance.  In particular, we have the following formulations in Eq.~\ref{eq:ahr}.
\begin{align}
	\label{eq:ahr}
	&\mathrm{Acc.}=\frac{\mathrm{TP}+\mathrm{TN}}{\mathrm{TP}+\mathrm{TN}+\mathrm{FP}+\mathrm{FN}}; \\ \nonumber
	&\mathrm{HR}=\frac{\mathrm{TP}}{\mathrm{TP}+\mathrm{FN}},
\end{align}

\noindent
where TP indicates trure positive, TN represents true negative, FP is false positive, and FN is false negative.

\begin{table*}[ht]
	\caption{{Overall Accuracy on Travel Time Estimation and Ranking Score Estimation}}
	\centering
	\begin{tabular}{l|lllllllll}
		\toprule[2pt]
		\multirow{3}{*}{\textbf{Method}} & \multicolumn{9}{l}{\textbf{Travel Time Estimation}}                                                                                                                                                                                                                                                                          \\ \cline{2-10} 
		& \multicolumn{3}{l|}{\textbf{Aalborg}}                                                                          & \multicolumn{3}{l|}{\textbf{Harbin}}                                                                            & \multicolumn{3}{l}{\textbf{Chengdu}}                                                    \\ \cline{2-10} 
		& \multicolumn{1}{l}{\textbf{MAE}}   & \multicolumn{1}{l}{\textbf{MARE}} & \multicolumn{1}{l|}{\textbf{MAPE}}  & \multicolumn{1}{l}{\textbf{MAE}}    & \multicolumn{1}{l}{\textbf{MARE}} & \multicolumn{1}{l|}{\textbf{MAPE}}  & \multicolumn{1}{l}{\textbf{MAE}}    & \multicolumn{1}{l}{\textbf{MARE}} & \textbf{MAPE}  \\ \toprule[1pt]
		\emph{Node2vec}                         & \multicolumn{1}{l}{63.82}          & \multicolumn{1}{l}{0.28}          & \multicolumn{1}{l|}{45.67}          & \multicolumn{1}{l}{269.21}          & \multicolumn{1}{l}{0.28}          & \multicolumn{1}{l|}{31.41}          & \multicolumn{1}{l}{290.47}          & \multicolumn{1}{l}{0.31}          & 34.43          \\ \hline
		\emph{DGI}                              & \multicolumn{1}{l}{67.22}          & \multicolumn{1}{l}{0.30}          & \multicolumn{1}{l|}{49.36}          & \multicolumn{1}{l}{288.09}          & \multicolumn{1}{l}{0.30}          & \multicolumn{1}{l|}{34.01}          & \multicolumn{1}{l}{312.28}          & \multicolumn{1}{l}{0.33}          & 38.46          \\ \hline
		\emph{GMI}                              & \multicolumn{1}{l}{70.61}          & \multicolumn{1}{l}{0.31}          & \multicolumn{1}{l|}{52.40}          & \multicolumn{1}{l}{310.39}          & \multicolumn{1}{l}{0.32}          & \multicolumn{1}{l|}{36.60}          & \multicolumn{1}{l}{337.06}          & \multicolumn{1}{l}{0.36}          & 41.58          \\ \hline
		\emph{MB}                               & \multicolumn{1}{l}{57.32}          & \multicolumn{1}{l}{0.25}          & \multicolumn{1}{l|}{39.37}          & \multicolumn{1}{l}{315.25}          & \multicolumn{1}{l}{0.31}          & \multicolumn{1}{l|}{35.28}          & \multicolumn{1}{l}{333.73}                & \multicolumn{1}{l}{0.35}              &42.45                \\ \hline
		\emph{BERT}                             & \multicolumn{1}{l}{71.96}          & \multicolumn{1}{l}{0.32}          & \multicolumn{1}{l|}{45.42}          & \multicolumn{1}{l}{217.96}          & \multicolumn{1}{l}{0.22}          & \multicolumn{1}{l|}{24.52}          & \multicolumn{1}{l}{303.00}                & \multicolumn{1}{l}{0.32}              &36.77                \\ \hline
		\emph{InfoGraph}                        & \multicolumn{1}{l}{69.36}          & \multicolumn{1}{l}{0.30}          & \multicolumn{1}{l|}{41.28}          & \multicolumn{1}{l}{200.81}          & \multicolumn{1}{l}{0.21}          & \multicolumn{1}{l|}{22.68}          & \multicolumn{1}{l}{291.54}          & \multicolumn{1}{l}{0.31}          & 36.07          \\ \hline
		\emph{PIM}                              & \multicolumn{1}{l}{57.66}          & \multicolumn{1}{l}{0.31}          & \multicolumn{1}{l|}{39.34}          & \multicolumn{1}{l}{196.06}          & \multicolumn{1}{l}{0.21}          & \multicolumn{1}{l|}{21.96}          & \multicolumn{1}{l}{289.10}          & \multicolumn{1}{l}{0.31}          & 35.55          \\ \hline
		\emph{DeepGTT}                          & \multicolumn{1}{l}{44.78}          & \multicolumn{1}{l}{0.20}          & \multicolumn{1}{l|}{26.53}          & \multicolumn{1}{l}{214.95}          & \multicolumn{1}{l}{0.22}          & \multicolumn{1}{l|}{22.76}          & \multicolumn{1}{l}{305.08}                & \multicolumn{1}{l}{0.33}              &35.47                \\ \hline
		\emph{HMTRL}                            & \multicolumn{1}{l}{40.59}          & \multicolumn{1}{l}{0.18}          & \multicolumn{1}{l|}{21.81}          & \multicolumn{1}{l}{228.58}          & \multicolumn{1}{l}{0.24}          & \multicolumn{1}{l|}{23.60}          & \multicolumn{1}{l}{360.08}                & \multicolumn{1}{l}{0.38}              &37.33                \\ \hline
		\emph{PathRank}                         & \multicolumn{1}{l}{37.09}          & \multicolumn{1}{l}{0.16}          & \multicolumn{1}{l|}{23.89}          & \multicolumn{1}{l}{190.08}          & \multicolumn{1}{l}{0.20}          & \multicolumn{1}{l|}{20.12}          & \multicolumn{1}{l}{334.94}          & \multicolumn{1}{l}{0.32}          & 35.11          \\ \hline
		\emph{GCN}                              & \multicolumn{1}{l}{78.04}               & \multicolumn{1}{l}{0.34}              & \multicolumn{1}{l|}{53.05}               & \multicolumn{1}{l}{368.21}          & \multicolumn{1}{l}{0.35}          & \multicolumn{1}{l|}{35.62}          & \multicolumn{1}{l}{480.83}                & \multicolumn{1}{l}{0.48}              &42.01                \\ \hline
		\emph{STGCN}                            & \multicolumn{1}{l}{58.57}               & \multicolumn{1}{l}{0.26}              & \multicolumn{1}{l|}{38.97}               & \multicolumn{1}{l}{284.12}                & \multicolumn{1}{l}{0.27}              & \multicolumn{1}{l|}{23.48}               & \multicolumn{1}{l}{406.09}                & \multicolumn{1}{l}{0.41}              &33.58                \\ \hline
		\emph{WSCCL}                            & \multicolumn{1}{l}{\textbf{31.66}} & \multicolumn{1}{l}{\textbf{0.14}} & \multicolumn{1}{l|}{\textbf{21.39}} & \multicolumn{1}{l}{\textbf{178.89}} & \multicolumn{1}{l}{\textbf{0.18}} & \multicolumn{1}{l|}{\textbf{19.43}} & \multicolumn{1}{l}{\textbf{281.20}} & \multicolumn{1}{l}{\textbf{0.29}} & \textbf{33.30} \\ \toprule[2pt]
		\multirow{3}{*}{\textbf{Method}} & \multicolumn{9}{l}{\textbf{Path Ranking Estimation}}                                                                                                                                                                                                                                                                         \\ \cline{2-10} 
		& \multicolumn{3}{l|}{\textbf{Aalborg}}                                                                          & \multicolumn{3}{l|}{\textbf{Harbin}}                                                                            & \multicolumn{3}{l}{\textbf{Chengdu}}                                                    \\ \cline{2-10} 
		& \multicolumn{1}{l}{\textbf{MAE}}   & \multicolumn{1}{l}{\textbf{$\tau$}}  & \multicolumn{1}{l|}{\textbf{$\rho$}}   & \multicolumn{1}{l}{\textbf{MAE}}    & \multicolumn{1}{l}{\textbf{$\tau$}}  & \multicolumn{1}{l|}{\textbf{$\rho$}}   & \multicolumn{1}{l}{\textbf{MAE}}    & \multicolumn{1}{l}{\textbf{$\tau$}}  & \textbf{$\rho$}   \\ \toprule[1pt]
		\emph{Node2vec}                         & \multicolumn{1}{l}{0.23}           & \multicolumn{1}{l}{0.60}          & \multicolumn{1}{l|}{0.64}           & \multicolumn{1}{l}{0.22}            & \multicolumn{1}{l}{0.37}          & \multicolumn{1}{l|}{0.40}           & \multicolumn{1}{l}{0.20}                & \multicolumn{1}{l}{0.73}              &0.75                \\ \hline
		\emph{DGI}                              & \multicolumn{1}{l}{0.24}           & \multicolumn{1}{l}{0.60}          & \multicolumn{1}{l|}{0.64}           & \multicolumn{1}{l}{0.21}            & \multicolumn{1}{l}{0.48}          & \multicolumn{1}{l|}{0.53}           & \multicolumn{1}{l}{0.21}                & \multicolumn{1}{l}{0.52}              &0.54                \\ \hline
		\emph{GMI}                              & \multicolumn{1}{l}{0.24}           & \multicolumn{1}{l}{0.59}          & \multicolumn{1}{l|}{0.62}           & \multicolumn{1}{l}{0.21}            & \multicolumn{1}{l}{0.49}          & \multicolumn{1}{l|}{0.54}           & \multicolumn{1}{l}{0.21}                & \multicolumn{1}{l}{0.51}              &0.53                \\ \hline
		\emph{MB}                               & \multicolumn{1}{l}{0.23}           & \multicolumn{1}{l}{0.62}          & \multicolumn{1}{l|}{0.66}           & \multicolumn{1}{l}{0.22}            & \multicolumn{1}{l}{0.44}          & \multicolumn{1}{l|}{0.48}           & \multicolumn{1}{l}{0.20}                & \multicolumn{1}{l}{0.71}              &0.74                \\ \hline
		\emph{BERT}                             & \multicolumn{1}{l}{0.26}           & \multicolumn{1}{l}{0.49}          & \multicolumn{1}{l|}{0.54}           & \multicolumn{1}{l}{0.22}            & \multicolumn{1}{l}{0.46}          & \multicolumn{1}{l|}{0.48}           & \multicolumn{1}{l}{0.22}                & \multicolumn{1}{l}{0.55}              &0.59                \\ \hline
		\emph{InfoGraph}                        & \multicolumn{1}{l}{0.26}           & \multicolumn{1}{l}{0.52}          & \multicolumn{1}{l|}{0.56}           & \multicolumn{1}{l}{0.21}            & \multicolumn{1}{l}{0.45}          & \multicolumn{1}{l|}{0.49}           & \multicolumn{1}{l}{0.20}                & \multicolumn{1}{l}{0.73}              &0.76                \\ \hline
		PIM                              & \multicolumn{1}{l}{0.22}           & \multicolumn{1}{l}{0.60}          & \multicolumn{1}{l|}{0.65}           & \multicolumn{1}{l}{0.21}            & \multicolumn{1}{l}{0.43}          & \multicolumn{1}{l|}{0.48}           & \multicolumn{1}{l}{0.19}                & \multicolumn{1}{l}{0.76}              &0.79                \\ \hline
		\emph{DeepGTT}                          & \multicolumn{1}{l}{0.39}           & \multicolumn{1}{l}{0.12}          & \multicolumn{1}{l|}{0.12}           & \multicolumn{1}{l}{0.29}            & \multicolumn{1}{l}{0.04}          & \multicolumn{1}{l|}{0.04}           & \multicolumn{1}{l}{0.23}                & \multicolumn{1}{l}{0.20}              &0.22                \\ \hline
		\emph{HMTRL}                            & \multicolumn{1}{l}{0.17}           & \multicolumn{1}{l}{0.65}          & \multicolumn{1}{l|}{0.68}           & \multicolumn{1}{l}{0.22}            & \multicolumn{1}{l}{0.51}          & \multicolumn{1}{l|}{0.56}           & \multicolumn{1}{l}{0.16}                & \multicolumn{1}{l}{0.77}              &0.79                \\ \hline
		\emph{PathRank}                         & \multicolumn{1}{l}{0.23}           & \multicolumn{1}{l}{0.64}          & \multicolumn{1}{l|}{0.68}           & \multicolumn{1}{l}{0.18}            & \multicolumn{1}{l}{0.55}          & \multicolumn{1}{l|}{0.60}           & \multicolumn{1}{l}{0.17}                & \multicolumn{1}{l}{0.79}              &0.83                \\ \hline
		\emph{WSCCL}                            & \multicolumn{1}{l}{\textbf{0.15}}  & \multicolumn{1}{l}{\textbf{0.68}} & \multicolumn{1}{l|}{\textbf{0.72}}  & \multicolumn{1}{l}{\textbf{0.14}}   & \multicolumn{1}{l}{\textbf{0.68}} & \multicolumn{1}{l|}{\textbf{0.73}}  & \multicolumn{1}{l}{\textbf{0.13}}                & \multicolumn{1}{l}{\textbf{0.84}}              &\textbf{0.86}                \\ \toprule[2pt]
	\end{tabular}
	\label{tb1:ttpr}

\end{table*}
\subsubsection{Baselines}
We compare \emph{WSCCL} with 12 baseline methods, which include 7 unsupervised methods and 5 supervised methods. The unsupervised methods are: 
\begin{itemize}
	\item  \textbf{Node2vec}~\cite{DBLP:conf/kdd/GroverL16}, Deep Graph InfoMax (\textbf{DGI})~\cite{DBLP:conf/iclr/VelickovicFHLBH19}, Graphical Mutual Information Maximization (\textbf{GMI})~\cite{DBLP:conf/www/PengHLZRXH20} are unsupervised graph representation learning frameworks, that give the edge representation for each edge in a graph. We use the average of the edge representations of the edge in a path as the path's representation. 
	\item \textbf{Memory Bank (\textbf{MB})}~\cite{DBLP:conf/cvpr/WuXYL18} is an unsupervised learning approach to learn representations based on contrastive loss, where the representations of negative sample are randomly selected from memory bank. We re-implement \textbf{MB} with an LSTM encoder to capture the sequential information in paths.
	\item \textbf{InfoGraph}~\cite{DBLP:conf/iclr/SunHV020} is a graph representation learning framework for unsupervised and semi-supervised settings. Here, we consider the unsupervised variant and treat a path as a graph to learn the path's representation. 
	\item \textbf{BERT}~\cite{DBLP:conf/naacl/DevlinCLT19} is an unsupervised language representation learning model. To enable training, we treat a path as a sentence and mask some edges in the path. Then we split a path $P$ into sub-paths $P_1$ and $P_2$, and consider $(P_1, P_2)$ as a valid question-answer~(Q\&A) pair and $(P_2, P_1)$ as an invalid Q\&A pair because the former represents a meaningful ordering while the latter does not. 
	\item \textbf{PIM}~\cite{DBLP:conf/ijcai/YangGHT021} is an unsupervised path representation learning model based on global and local mutual information maximization.
\end{itemize}
The supervised methods that take into account the labels from a specific downstream task to obtain path representations, which are:
\begin{itemize}
	\item \textbf{DeepGTT}~\cite{DBLP:conf/www/LiCSC19} is a \emph{supervised} travel time distribution estimation~(i.e., to learn the parameters for inverse Gaussian distribution) framework based on a deep generative model. 
	\item \textbf{HMTRL}~\cite{DBLP:journals/pvldb/0026HFZL020} enables unified route representation learning and exploits both spatio-temporal dependencies in road networks and the semantic coherence of historical routes.
	\item \textbf{PathRank}~\cite{yang2020context} is a \emph{supervised} path representation learning model based on GRUs.
	\item \textbf{GCN}~\cite{DBLP:conf/nips/DefferrardBV16} is a graph covolutional neural network based method that estimates the travel times of all edges in a road network. The travel time of a path is then the sum of the travel times of the edges in the path. 
	\item \textbf{STGCN}~\cite{DBLP:conf/ijcai/YuYZ18} is a traffic prediction framework based on spatio-temporal graph convolutional networks. Similar to GCN, the travel time of a path is the sum of the predicted travel times of the edges in the path. 
\end{itemize}

Finally, note that for the path ranking task, we cannot simply aggregate the rankings of edges to obtain the rankings of paths. This also applies to the newly included path recommendation task. Thus, GCNs and STGCNs cannot work as baselines for the these two tasks.

\subsubsection{Models for Downstream Tasks}

For all unsupervised learning approaches, we first obtain a task-independent TPR and then apply a regression model to address different downstream tasks using task-specific labels. %
In the experiments, we use the ensemble model \textit{Gradient Boosting Regressor} (\emph{GBR}) %
to estimate travel time and ranking scores for paths as they are regression problems. In addition, we use the ensemble model \textit{Gradient Boosting Classifier} (\emph{GBC}) to make path recommendations, as they are classification problems.

\subsubsection{Weak Labels}
We consider two different types of weak labels, including peak/off-peak (POP) and traffic congestion indices (TCI). We use POP as default weak labels. We only conduct experiments on the Harbin and Chengdu data sets for TCI since we cannot obtain TCI for Aalborg, Denmark from Baidu Maps\footnote{\href{https://jiaotong.baidu.com/congestion/city/urbanrealtime/}{https://jiaotong.baidu.com/congestion/city/urbanrealtime/}}. 

\subsubsection{Implementation Settings}
We set embedding feature dimensions of RT, NoL, OW, and TS as $d_{rt}=64$, $d_l=32$, $d_o=16$, $d_{ts}=16$, respectively. The feature dimensions of output from \emph{Node2Vec} on both temporal graph and road networks are set to be 128. 
Meanwhile, we apply 2 LSTM layers and set the dimensionality of the hidden state $h_j$ to 128. Further, we set the size for temporal path representation dimensionality to 128, i.e., $d_h=128$. The number of Meta-Set is set to be $N=10$, and the number of stages in curriculum learning is also set to be $M=10$. 
The hyper-parameter $\lambda$ is set to 0.8. We set the learning rate (lr) to $3e-4$ and the batch size 32. 
In particular, 
we train our \emph{WSCCL} using all unlabeled paths shown in data sets section and then
we randomly choose 80\% and 20\% paths in labeled path as training and testing data for GBR. Finally, we evaluate all models on a powerful Linux server with 40 Intel(R) Xeon(R) Gold 5215 CPUs @ 2.50GHz and four Quadro RTX 8000 GPU cards. Finally, all algorithm are implemented in PyTorch 1.9.1. The code is available at \href{https://github.com/Sean-Bin-Yang/TPR.git}{https://github.com/Sean-Bin-Yang/TPR.git}.

\subsection{Experimental Results}

\subsubsection{\textbf{Overall accuracy on downstream tasks}}

Table~\ref{tb1:ttpr}, and Table~\ref{tb:hit} report the overall results on the three downstream tasks.
\emph{WSCCL} achieves the best performance on these three tasks for three real-world data sets.
The three graph node representation learning methods \emph{Node2vec}, \emph{DGI}, and \emph{GMI} are unable to capture temporal correlation in the temporal path. In contrast, \emph{WSCCL} takes temporal correlation into consideration by virtue of its temporal embedding layer. In addition, the weakly supervised contrastive curriculum learning improves the estimation accuracy.

\begin{table}[ht]
\caption{Overall Performance on Path
Recommendation}
\centering
\begin{tabular}{l|ll|ll|ll}
\toprule[2pt]
\multirow{2}{*}{\textbf{Methods}}          & \multicolumn{2}{l|}{\textbf{Aalborg}}          & \multicolumn{2}{l|}{\textbf{Harbin}}           & \multicolumn{2}{l}{\textbf{Chengdu}}          \\ \cline{2-7}
          & \multicolumn{1}{l}{\textbf{Acc.}} & \textbf{HR} & \multicolumn{1}{l}{\textbf{Acc.}} & \textbf{HR} & \multicolumn{1}{l}{\textbf{Acc.}} & \textbf{HR} \\ \hline
\emph{Node2vec}  & \multicolumn{1}{l}{0.79}         &0.51       & \multicolumn{1}{l}{0.76}         &0.51       & \multicolumn{1}{l}{0.75}         &0.61       \\ \hline
\emph{DGI}       & \multicolumn{1}{l}{0.74}         &0.55       & \multicolumn{1}{l}{0.70}         &0.36       & \multicolumn{1}{l}{0.70}         &0.57       \\ \hline
\emph{GMI}       & \multicolumn{1}{l}{0.78}         &0.53       & \multicolumn{1}{l}{0.72}         &0.41       & \multicolumn{1}{l}{0.68}         &0.58       \\ \hline
\emph{MB}        & \multicolumn{1}{l}{0.67}         &0.48       & \multicolumn{1}{l}{0.61}         &0.69       & \multicolumn{1}{l}{0.73}         &0.69       \\ \hline
\emph{BERT}      & \multicolumn{1}{l}{0.60}         &0.43       & \multicolumn{1}{l}{0.64}         &0.53       & \multicolumn{1}{l}{0.66}         &0.61       \\ \hline
\emph{InfoGraph} & \multicolumn{1}{l}{0.72}         &0.69      & \multicolumn{1}{l}{0.79}         &0.78       & \multicolumn{1}{l}{0.73}         &0.65       \\ \hline
\emph{PIM}       & \multicolumn{1}{l}{0.79}         &0.82       & \multicolumn{1}{l}{0.86}         &0.83       & \multicolumn{1}{l}{0.76}         &0.74       \\ \hline
\emph{HMTRL}     & \multicolumn{1}{l}{0.80}         &0.86       & \multicolumn{1}{l}{0.81}         &0.82       & \multicolumn{1}{l}{0.78}         &0.83       \\ \hline
\emph{PathRank}  & \multicolumn{1}{l}{0.77}         &0.71       & \multicolumn{1}{l}{0.79}         &0.74       & \multicolumn{1}{l}{0.77}         &0.73       \\ \hline
\emph{WSCCL}     & \multicolumn{1}{l}{\textbf{0.82}}         &\textbf{0.88}       & \multicolumn{1}{l}{\textbf{0.97}}         &\textbf{0.91}       & \multicolumn{1}{l}{\textbf{0.81}}         &\textbf{0.90}      \\ \toprule[2pt]
\end{tabular}
\label{tb:hit}

\end{table}
Although \emph{MB} and \emph{BERT} can capture dependencies among the edge feature vectors in temporal paths, these approaches achieve poor estimation accuracy. This is because \emph{MB} needs large amounts of negative samples to ensure effective training, which is not feasible in our scenario. In addition, \emph{BERT} is not well suited for our setting of learning generic TPRs since BERT cannot support contrastive learning in the setting of multiple positive samples against multiple negative samples. 

\emph{InfoGraph} learns full graph representations. However, it only applies a local view and cannot capture the sequential information of edge in a path. In contrast, \emph{WSCCL} not only uses sequence model (e.g., LSTM) to capture sequential information between edge in a path but considers both the (global) path and (local) edge levels.  
Although \emph{PIM} is designed for path representation learning, it is unable to learn meaningful TPRs because it ignores temporal information and only has one positive sample.  In contrast, \emph{WSCCL} allows multiple positive temporal path samples in each minibatch and also uses a learned curriculum instead of the pre-defined curriculum negative sampling used by \emph{PIM}.
The supervised learning methods \emph{DeepGTT}, \emph{HMTRL}, and \emph{PathRank} achieve relatively poor accuracy due to the small size of labeled training data. Since task-specific labels~(``strong labels'') are expensive to obtain, we consider a setting where labelled training data is limited. \emph{DeepGTT} exhibits the worst performance on the Path Ranking task because it is designed for travel-time estimation, which is evidence for the poor generalizability of supervised feature representation learning, as discussed in Section~\ref{sec:intro}. In contrast, the \emph{GCN} and \emph{STGCN} results also are worse than the \emph{WSCCL} results. This is because dependencies among edges are disregarded.

\begin{figure}[t]
\centering    
\subfigure[Travel Time (Aalborg)] {
 \label{fig:a}     
\includegraphics[height=3.7cm, width=0.44\columnwidth]{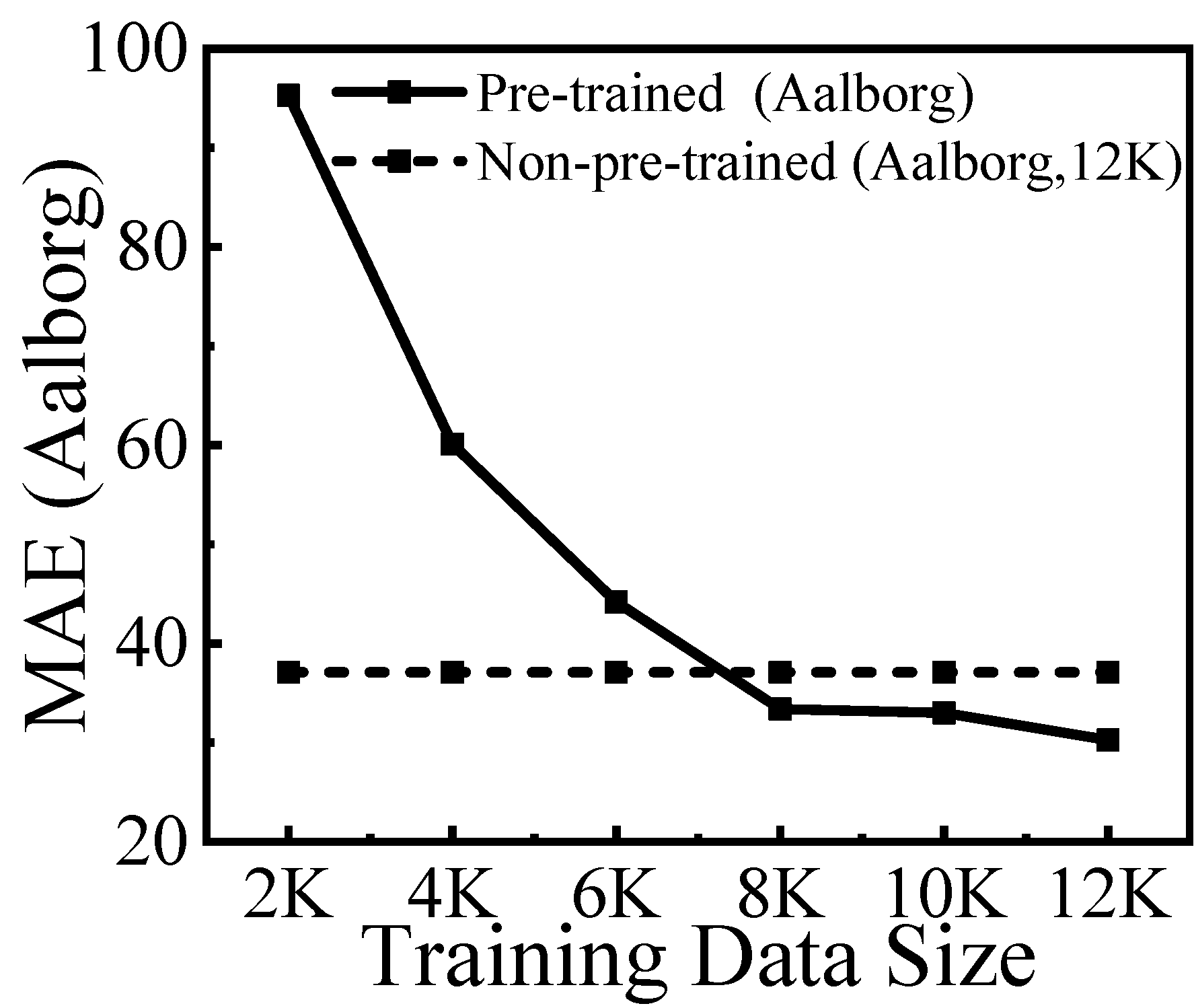}  
}  
\vspace{-5pt}
\subfigure[Path Ranking (Aalborg)] { 
\label{fig:b}     
\includegraphics[height=3.6cm, width=0.44\columnwidth]{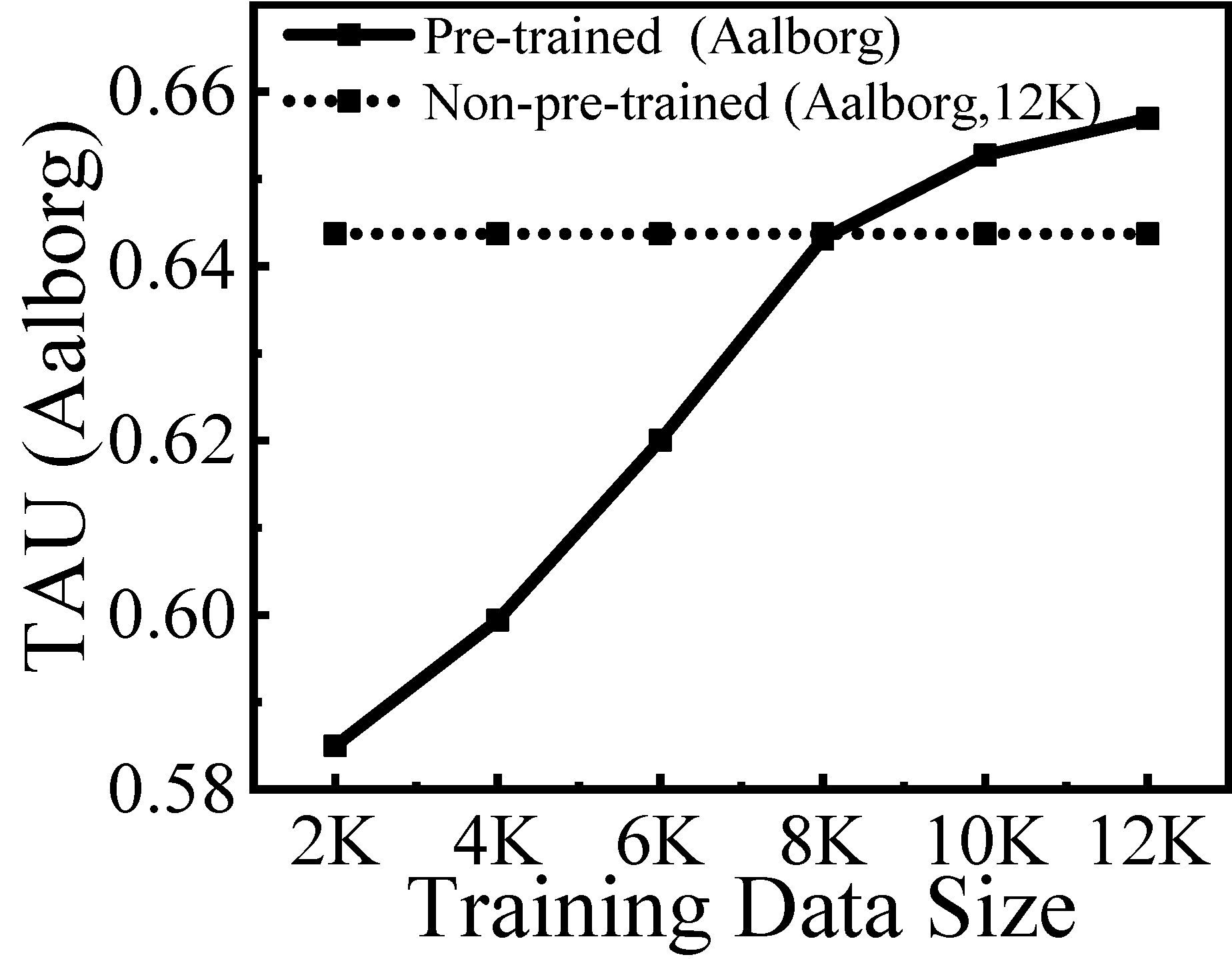}     
}   
\vspace{-5pt}
\subfigure[Travel Time (Harbin)] {
 \label{fig:c}     
\includegraphics[height=3.7cm, width=0.44\columnwidth]{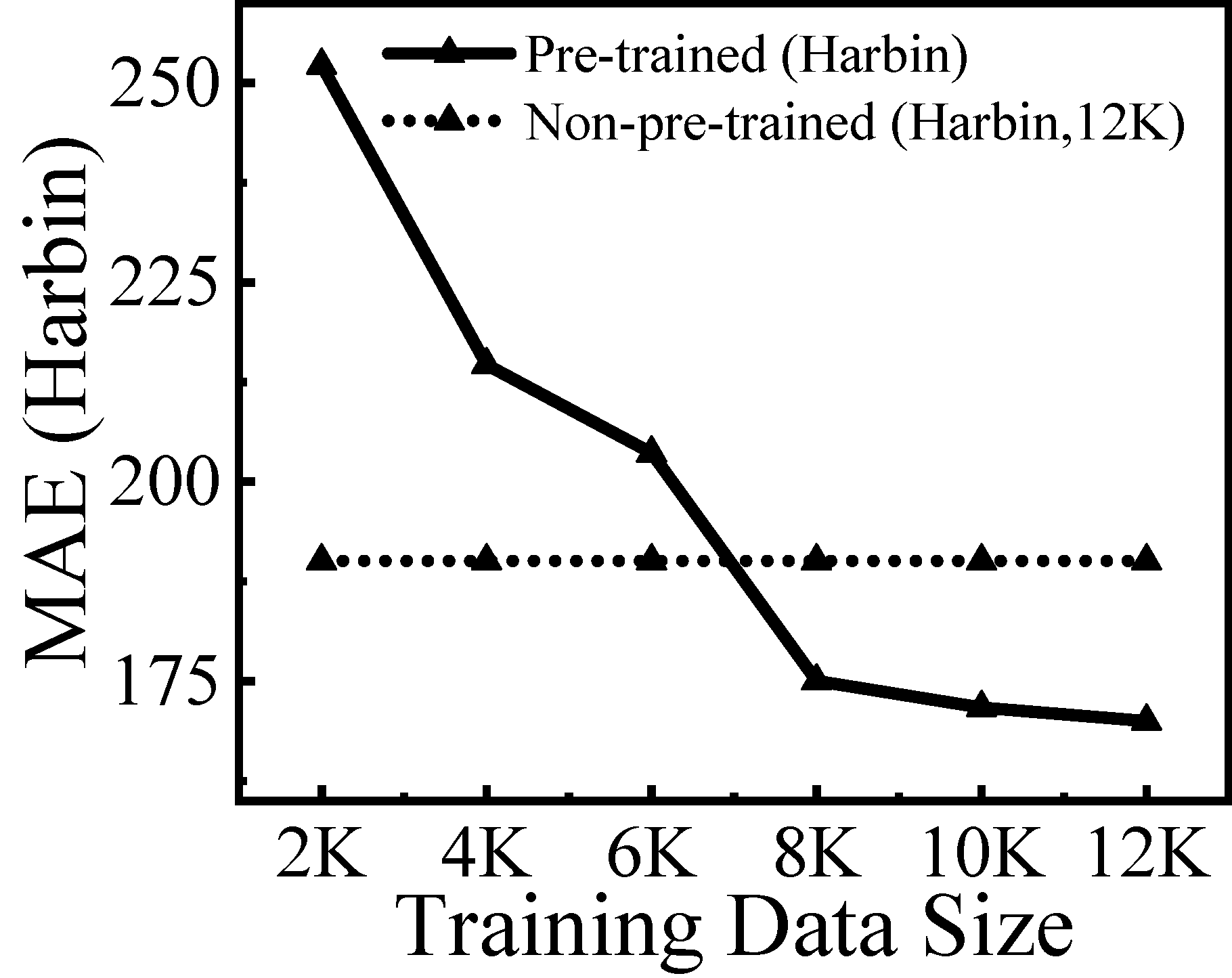}  
}  
\subfigure[Path Ranking (Harbin)] { 
\label{fig:d}     
\includegraphics[height=3.7cm, width=0.44\columnwidth]{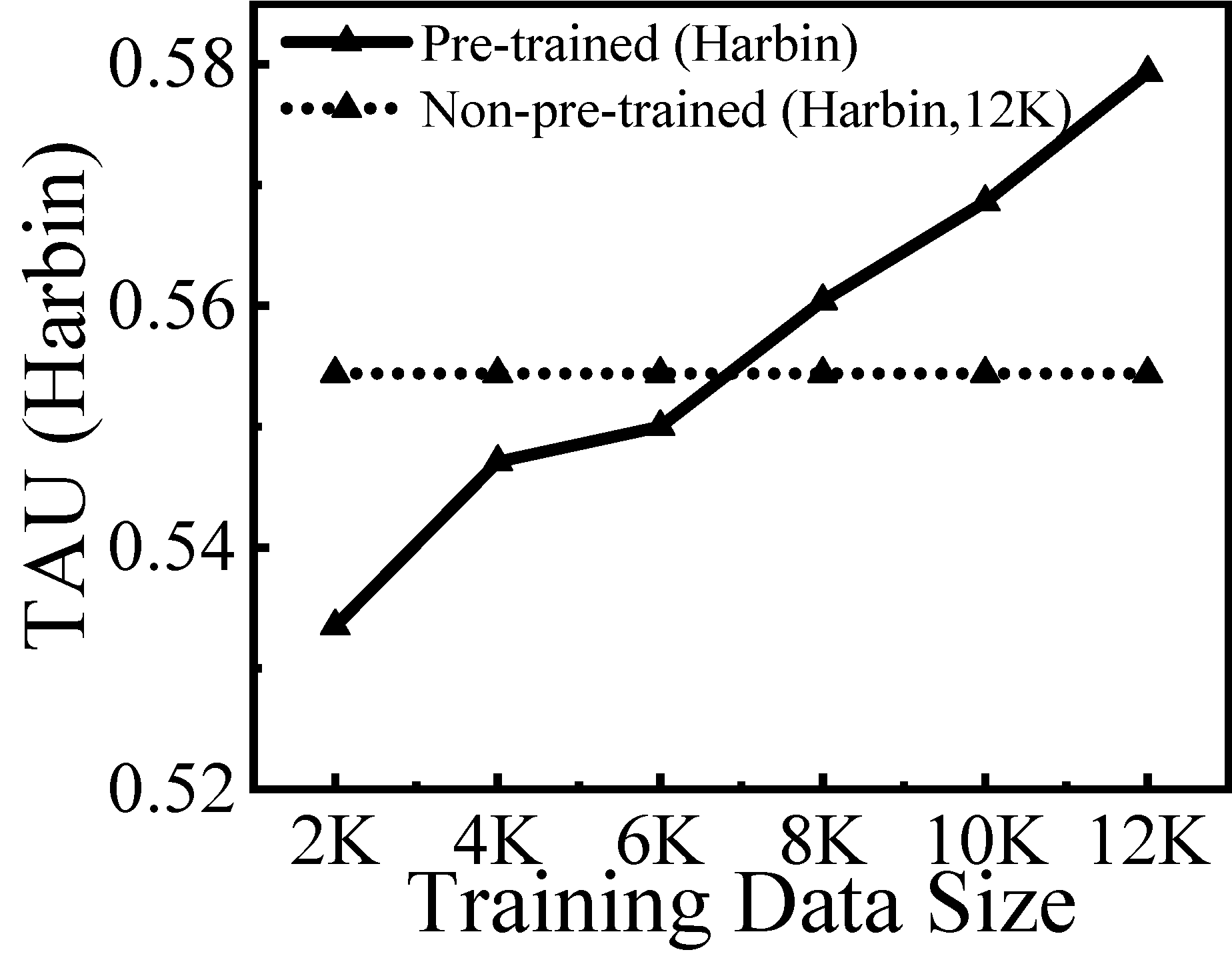}     
}

\subfigure[Travel Time (Chengdu)] {
 \label{fig:e}     
\includegraphics[height=3.7cm, width=0.44\columnwidth]{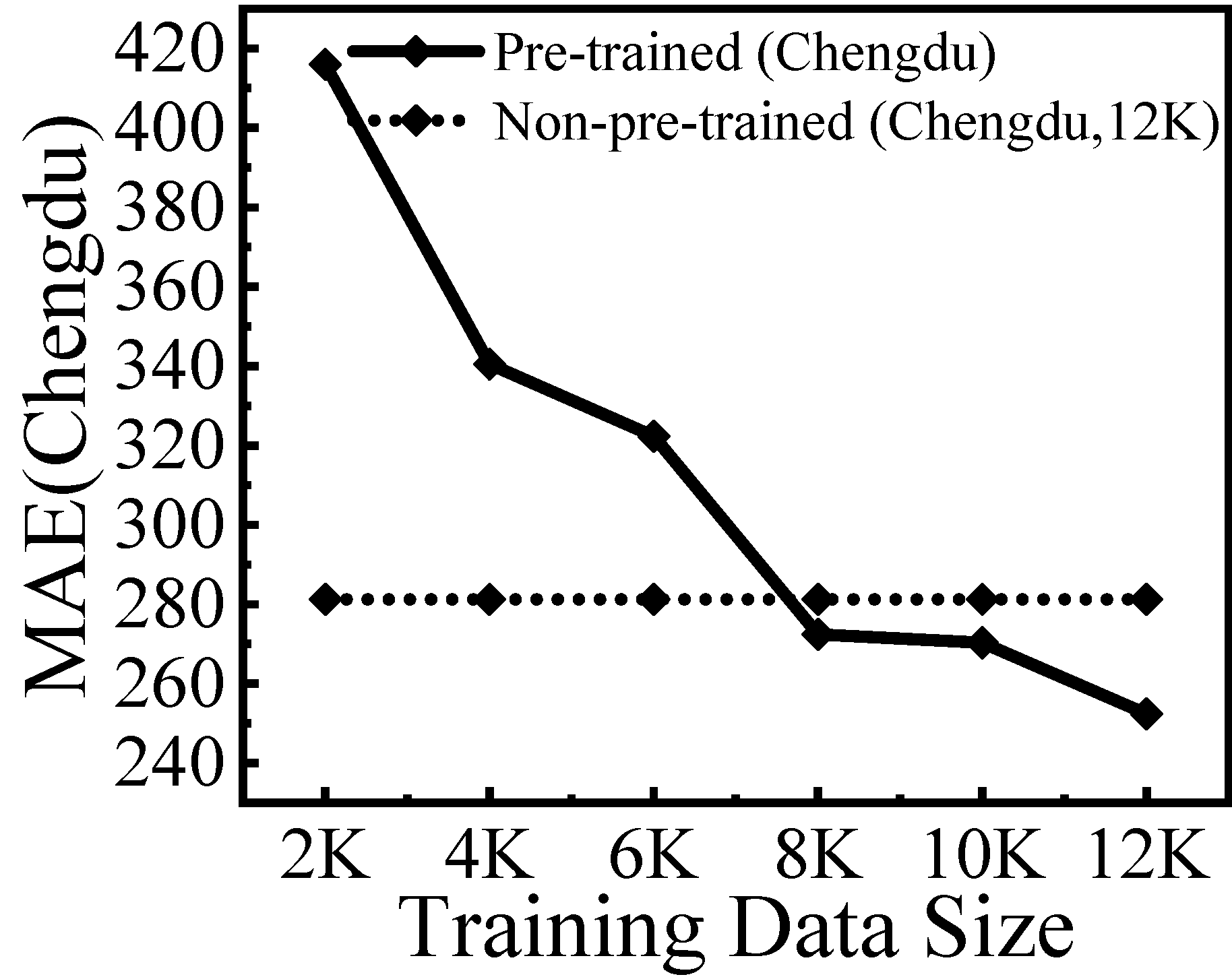}  
} 
\vspace{-5pt}
\subfigure[Path Ranking (Chengdu)] { 
\label{fig:f}     
\includegraphics[height=3.7cm, width=0.44\columnwidth]{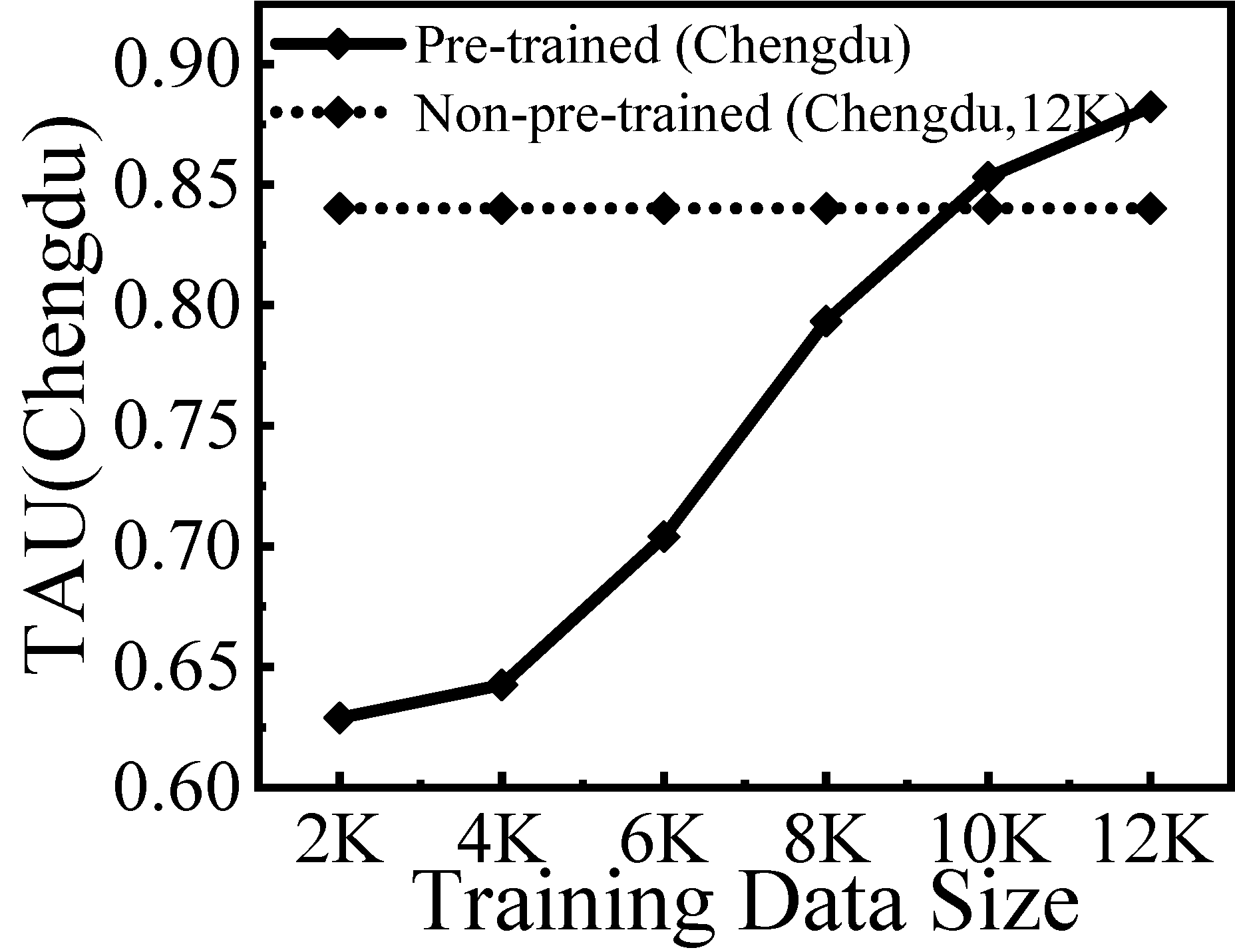}     
}
\caption{Effects of Pre-training}  
\label{fig:pre-tt-pr}     

\end{figure}
\subsubsection{\textbf{Using \emph{WSCCL} as a Pre-training Method}}

We conduct experiments that treat \emph{WSCCL} as a pre-training method for the supervised method \emph{PathRank}. Here, \emph{PathRank} takes as input a sequence of edge features and estimates the travel time and ranking score. To use \emph{WSCCL} for pre-training method for \emph{PathRank}, we first train \emph{WSCCL} in a weakly supervised manner and then apply the learned parameters in temporal path encoder to initialize the encoder in \emph{PathRank}. 

Fig.~\ref{fig:pre-tt-pr} shows performance of \emph{PathRank} with and without pre-training for the two tasks. Without pre-training,  \emph{PathRank} is trained by using 12K labeled training paths. We observe the following: 1) With pre-training we can achieve the same performance as with non-pre-trained \emph{PathRank} while using fewer labeled training paths. For example, when using \emph{WSCCL} for pre-training, \emph{PathRank} only needs 8K, 7K and 10K labeled samples for the Aalborg, Harbin and Chengdu data sets, respectively, to achieve the same performance as \emph{PathRank} with 12K labeled samples, on the path ranking task. 2) When we pre-train \emph{PathRank} with 12K samples, the performance is much better than without pre-training. In both tasks, we obtain similar observations, which indicates that \emph{WSCCL} can be applied advantageously for the pre-training of different downstream tasks.

\subsubsection{\textbf{Ablation Studies}}
We conduct ablation studies on \emph{WSCCL} to observe 1)~the effect of the CL design strategy on TPR learning; 2)~the effect of variants of \emph{WSCCL}, specifically \emph{CL}, \emph{global loss}, and local loss; 3)~the effect of different weak labels; and 4)~the effect of temporal information.

\noindent
\paragraph{Effect of the CL Design Strategy}~To observe the effectiveness of the learned CL, we compare with \emph{WSCCL} with a heuristic curriculum design where we simply sort the paths based on the number of edges. The comparison between these two CL variants is reported in Table~\ref{tb:heu} that shows that our learned curriculum is better than the heuristic curriculum design on all tasks over on three data sets. This is because the lengths of edges are varying~(from few meters to kilometers),  even two paths with the same number of edges, the lengths of two paths may have a big difference. Thus, the difficulties of paths cannot simply represented by the number of edges directly. 

\begin{table}[tp]
\caption{{Effect of the CL Design Strategy}}
\centering
\begin{tabular}{l|llllll}
\toprule[2pt]
\multirow{3}{*}{\textbf{Methods}} & \multicolumn{6}{l}{\textbf{Aalborg}}                                                    \\ \cline{2-7} 
\multirow{2}{*}{} & \multicolumn{3}{l|}{\textbf{Travel Time Estimation}} & \multicolumn{3}{l}{\textbf{Path Ranking}} \\ \cline{2-7}
                  &\textbf{MAE} &\textbf{MARE} &\multicolumn{1}{l|}{\textbf{MAPE}} & \textbf{MAE}    &\textbf{$\tau$} &\textbf{$\rho$} \\ \toprule[1pt]
\textbf{Heuristic}  &33.58     &0.15    &\multicolumn{1}{l|}{22.06}       &0.19       &0.61           &0.65                \\ \hline
\textbf{WSCCL}   &\textbf{31.66}     &\textbf{0.13}      &\multicolumn{1}{l|}{\textbf{21.39}}      &\textbf{0.15}  &\textbf{0.68}     &\textbf{0.72}       \\ \toprule[2pt] 
\multirow{3}{*}{\textbf{Methods}} & \multicolumn{6}{l}{\textbf{Harbin}}                                                    \\ \cline{2-7} 
\multirow{2}{*}{} & \multicolumn{3}{l|}{\textbf{Travel Time Estimation}} & \multicolumn{3}{l}{\textbf{Path Ranking}} \\ \cline{2-7}
&\textbf{MAE} &\textbf{MARE} &\multicolumn{1}{l|}{\textbf{MAPE}} & \textbf{MAE}    &\textbf{$\tau$} &\textbf{$\rho$} \\ \toprule[1pt]
\textbf{Heuristic} &193.94      &0.18   &\multicolumn{1}{l|}{21.43}          &0.16       &0.58                 &0.62                 \\ \hline
\textbf{WSCCL}    &\textbf{178.89}     &\textbf{0.18}   &\multicolumn{1}{l|}{\textbf{19.43}}          &\textbf{0.14}      &\textbf{0.68}                &\textbf{0.73}                  \\ \toprule[2pt] 
\multirow{3}{*}{\textbf{Methods}} & \multicolumn{6}{l}{\textbf{ChengDu}}                                                    \\ \cline{2-7} 
\multirow{2}{*}{} & \multicolumn{3}{l|}{\textbf{Travel Time Estimation}} & \multicolumn{3}{l}{\textbf{Path Ranking}} \\ \cline{2-7}
&\textbf{MAE} &\textbf{MARE} &\multicolumn{1}{l|}{\textbf{MAPE} }& \textbf{MAE}    &\textbf{$\tau$} &\textbf{$\rho$} \\ \toprule[1pt]
\textbf{Heuristic} &292.52      &0.31   &\multicolumn{1}{l|}{34.52}         &0.19       &0.64              &0.67                 \\ \hline
\textbf{WSCCL}    &\textbf{281.20} &\textbf{0.29} &\multicolumn{1}{l|}{\textbf{33.30}}          &\textbf{0.13}      &\textbf{0.84}                &\textbf{0.86}           \\ \toprule[2pt]
\end{tabular}

\label{tb:heu}
\end{table}

\begin{table}[tp]
\caption{Effects of CL, Global Loss and Local Loss}
\centering

\begin{tabular}{l|llllll}
\toprule[2pt]
\multirow{3}{*}{\textbf{Methods}} & \multicolumn{6}{l}{\textbf{Aalborg}}                                                    \\ \cline{2-7} 
\multirow{2}{*}{} & \multicolumn{3}{l|}{\textbf{Travel Time Estimation}} & \multicolumn{3}{l}{\textbf{Path Ranking}} \\ \cline{2-7}
                  &\textbf{MAE} &\textbf{MARE} &\multicolumn{1}{l|}{\textbf{MAPE}} & \textbf{MAE}    &\textbf{$\tau$} &\textbf{$\rho$} \\ \toprule[1pt]
\textbf{w/o CL}  &32.58    &0.14    &\multicolumn{1}{l|}{21.88}       &0.19       &0.62         &0.66           \\ \hline
\textbf{w/o Global}   &51.19    &0.22    &\multicolumn{1}{l|}{31.13}       &0.24       &0.54         &0.58         \\ \hline

\textbf{w/o Local}  &32.80    &0.14    &\multicolumn{1}{l|}{23.34}       &0.21       &0.57         &0.62       \\ \hline
\textbf{WSCCL}   &\textbf{31.66}     &\textbf{0.14}      &\multicolumn{1}{l|}{\textbf{21.39}}     &\textbf{0.15}  &\textbf{0.68}     &\textbf{0.72}       \\ \toprule[2pt] 

\multirow{3}{*}{\textbf{Methods}} & \multicolumn{6}{l}{\textbf{Harbin}}                                                    \\ \cline{2-7} 
\multirow{2}{*}{} & \multicolumn{3}{l|}{\textbf{Travel Time Estimation}} & \multicolumn{3}{l}{\textbf{Path Ranking}} \\ \cline{2-7}
&\textbf{MAE} &\textbf{MARE} &\multicolumn{1}{l|}{\textbf{MAPE}} & \textbf{MAE}    &\textbf{$\tau$} &\textbf{$\rho$} \\ \toprule[1pt]
\textbf{w/o CL} &184.52       &0.19    &\multicolumn{1}{l|}{20.44}           &0.18       &0.53                 &0.57                 \\ \hline
\textbf{w/o Global}      &200.76       &0.21    &\multicolumn{1}{l|}{22.91}           &0.22       &0.34                 &0.38                \\ \hline
\textbf{w/o Local}      &196.35       &0.20    &\multicolumn{1}{l|}{21.55}           &0.15       &0.65                 &0.69                 \\ \hline
\textbf{WSCCL}    &\textbf{178.89}     &\textbf{0.18}   &\multicolumn{1}{l|}{\textbf{19.43}}          &\textbf{0.14}      &\textbf{0.68}                &\textbf{0.73}                 \\ \toprule[2pt]

\multirow{3}{*}{\textbf{Methods}} & \multicolumn{6}{l}{\textbf{ChengDu}}                                                    \\ \cline{2-7} 
\multirow{2}{*}{} & \multicolumn{3}{l|}{\textbf{Travel Time Estimation}} & \multicolumn{3}{l}{\textbf{Path Ranking}} \\ \cline{2-7}
&\textbf{MAE} &\textbf{MARE} &\multicolumn{1}{l|}{\textbf{MAPE}} & \textbf{MAE}    &\textbf{$\tau$} &\textbf{$\rho$} \\ \toprule[1pt]
\textbf{w/o CL} &282.85       &0.30    &\multicolumn{1}{l|}{33.87}           &0.13       &0.83                 &0.85                \\ \hline
\textbf{w/o Global}      &299.45       &0.32    &\multicolumn{1}{l|}{36.86}           &0.18       &0.67                 &0.71               \\ \hline
\textbf{w/o Local}      &287.96       &0.31    &\multicolumn{1}{l|}{34.78}           &0.17      &0.68                 &0.72                 \\ \hline
\textbf{WSCCL}    &\textbf{281.20} &\textbf{0.29} &\multicolumn{1}{l|}{\textbf{33.30}}          &\textbf{0.13}      &\textbf{0.84}                &\textbf{0.86}                \\ \toprule[2pt]
\end{tabular}

\label{tb:joint}
\end{table}

\noindent
\paragraph{Effects of Global Loss, Local Loss, and CL} To study the effect of these three modules, we consider three variants of \emph{WSCCL}: 1)~\emph{
w/o Global}, 2)~\emph{w/o Local}, and 3)~\emph{w/o CL}. In \emph{w/o Global}, we remove Global WSC loss from \emph{WSCCL}, in \emph{WSCCL} \emph{w/o Local}, we remove the Local WSC loss, and in \emph{WSCCL} \emph{w/o CL}, the curriculum strategy is omitted. The results on three data sets are reported in Table~\ref{tb:joint}. We can observe that \emph{WSCCL} \emph{w/o Global} shows the worst performance and has a clear margin to the other variants. This shows that the proposed Global WSC performs well. We also observe that \emph{WSCCL} achieves the best performance. This indicates that all the proposed modules contribute positively to the final performance, which validates the overall design.

\noindent
\paragraph{Effect of Different Weak Labels} We conduct additional experiments by using traffic congestion indices (TCI), which indicate four congestion levels in a city across time, as weak labels. 
Table~\ref{tb:tci} shows the results on the Harbin and Chengdu data sets. We observe that \emph{WSCCL} works well when using the TCI as weak labels. 

\noindent
\paragraph{Effect of Temporal Information} We further conduct experiments on the three data sets using a \emph{WSCCL} variant that disregards temporal information. The results, shown in Table~\ref{tab:nti}, indicate that the non-temporal \emph{WSCCL-NT} performs worse than \emph{WSCCL} on both downstream tasks, suggesting that our temporal embedding is effective.

\begin{table}[tp]
\caption{Effect of Different Weak Labels}
\centering
\begin{tabular}{l|llllll}
\toprule[2pt]
\multirow{3}{*}{\textbf{Methods}} & \multicolumn{6}{l}{\textbf{Harbin}}                                                    \\ \cline{2-7} 
\multirow{2}{*}{} & \multicolumn{3}{l|}{\textbf{Travel Time Estimation}} & \multicolumn{3}{l}{\textbf{Path Ranking}} \\ \cline{2-7}
                  &\textbf{MAE} &\textbf{MARE} &\multicolumn{1}{l|}{\textbf{MAPE}} & \textbf{MAE}    &\textbf{$\tau$} &\textbf{$\rho$} \\ \toprule[1pt]
\emph{WSCCL-TCI}  &177.07     &0.18    &\multicolumn{1}{l|}{19.19}      &0.13        &0.70           &0.74              \\ \hline
\emph{WSCCL-POP}   &178.89    &0.18      &\multicolumn{1}{l|}{19.43}     &0.14  &0.68     &0.73       \\ \toprule[2pt] 
\multirow{3}{*}{\textbf{Methods}} & \multicolumn{6}{l}{\textbf{Chengdu}}                                                    \\ \cline{2-7} 
\multirow{2}{*}{} & \multicolumn{3}{l|}{\textbf{Travel Time Estimation}} & \multicolumn{3}{l}{\textbf{Path Ranking}} \\ \cline{2-7}
                  &\textbf{MAE} &\textbf{MARE} &\multicolumn{1}{l|}{\textbf{MAPE}} & \textbf{MAE}    &\textbf{$\tau$} &\textbf{$\rho$} \\ \toprule[1pt]
\emph{WSCCL-TCI}  &280.85     &0.29    &\multicolumn{1}{l|}{32.97}       &0.12        &0.86           &0.87              \\ \hline
\emph{WSCCL-POP}   &281.20    &0.29      &\multicolumn{1}{l|}{33.30}     &0.13  &0.84     &0.86       \\ \toprule[2pt]
\end{tabular}

\label{tb:tci}
\end{table}

\begin{table}[tp]
\caption{Effect of Temporal Information}
\centering
\begin{tabular}{l|llllll}
\toprule[2pt]
\multirow{3}{*}{\textbf{Methods}} & \multicolumn{6}{l}{\textbf{Aalborg}}                                                                                                                 \\ \cline{2-7} 
                  & \multicolumn{3}{l|}{\textbf{Travel Time Estimation}}                                      & \multicolumn{3}{l}{\textbf{PathRank}}                             \\ \cline{2-7} 
                  & \multicolumn{1}{l}{\textbf{MAE}} & \multicolumn{1}{l}{\textbf{MARE}} & \multicolumn{1}{l|}{\textbf{MAPE}} & \multicolumn{1}{l}{\textbf{MAE}} & \multicolumn{1}{l}{$\tau$} & $\rho$ \\ \hline
\emph{WSCCL}             & \multicolumn{1}{l}{\textbf{31.66}}    & \multicolumn{1}{l}{\textbf{0.14}}     & \multicolumn{1}{l|}{\textbf{21.39}}     & \multicolumn{1}{l}{\textbf{0.15}}    & \multicolumn{1}{l}{\textbf{0.68}}    &\textbf{0.72}     \\ \hline
\emph{WSCCL-NT}  & \multicolumn{1}{l}{41.25}    & \multicolumn{1}{l}{0.18}     & \multicolumn{1}{l|}{29.38}     & \multicolumn{1}{l}{0.21}    & \multicolumn{1}{l}{0.55}    &0.59     \\ \toprule[2pt]
\multirow{3}{*}{\textbf{Methods}} & \multicolumn{6}{l}{\textbf{Harbin}}                                                                                                                 \\ \cline{2-7}
                  & \multicolumn{3}{l|}{\textbf{Travel Time Estimation}}                                      & \multicolumn{3}{l}{\textbf{PathRank}}                             \\ \cline{2-7}
                  & \multicolumn{1}{l}{\textbf{MAE}} & \multicolumn{1}{l}{\textbf{MARE}} & \multicolumn{1}{l|}{\textbf{MAPE}} & \multicolumn{1}{l}{\textbf{MAE}} & \multicolumn{1}{l}{$\tau$} & $\rho$ \\ \hline
\emph{WSCCL}             & \multicolumn{1}{l}{\textbf{178.89}}    & \multicolumn{1}{l}{\textbf{0.18}}     & \multicolumn{1}{l|}{\textbf{19.43}}     & \multicolumn{1}{l}{\textbf{0.14}}    & \multicolumn{1}{l}{\textbf{0.68}}    &\textbf{0.73}     \\ \hline
\emph{WSCCL-NT}  & \multicolumn{1}{l}{199.58}    & \multicolumn{1}{l}{0.20}     & \multicolumn{1}{l|}{22.20}     & \multicolumn{1}{l}{0.15}    & \multicolumn{1}{l}{0.64}    &0.68     \\ \toprule[2pt]
\multirow{3}{*}{\textbf{Methods}}                   & \multicolumn{6}{l}{\textbf{Chengdu}}                                                                                                                 \\ \cline{2-7}
                  & \multicolumn{3}{l|}{\textbf{Travel Time Estimation}}                                      & \multicolumn{3}{l}{\textbf{PathRank}}                             \\ \cline{2-7}
                  & \multicolumn{1}{l}{\textbf{MAE}} & \multicolumn{1}{l}{\textbf{MARE}} & \multicolumn{1}{l|}{\textbf{MAPE}} & \multicolumn{1}{l}{\textbf{MAE}} & \multicolumn{1}{l}{$\tau$} & $\rho$ \\ \hline
\emph{WSCCL}             & \multicolumn{1}{l}{\textbf{281.20}}    & \multicolumn{1}{l}{\textbf{0.29}}     & \multicolumn{1}{l|}{\textbf{33.30}}     & \multicolumn{1}{l}{\textbf{0.13}}    & \multicolumn{1}{l}{\textbf{0.84}}    &\textbf{0.86}     \\ \hline
\emph{WSCCL-NT}  & \multicolumn{1}{l}{292.76}    & \multicolumn{1}{l}{0.31}     & \multicolumn{1}{l|}{35.10}     & \multicolumn{1}{l}{0.18}    & \multicolumn{1}{l}{0.81}    &0.83     \\ \toprule[2pt]
\end{tabular}
\label{tab:nti}

\end{table}

\begin{table}[tp]
\caption{{Comparison with Temporally Enhanced Unsupervised \emph{PIM} Method}}
\centering

\begin{tabular}{l|llllll}
\toprule[2pt]
\multirow{3}{*}{\textbf{Methods}} & \multicolumn{6}{l}{\textbf{Aalborg}}                                                    \\ \cline{2-7} 
\multirow{2}{*}{} & \multicolumn{3}{l|}{\textbf{Travel Time Estimation}} & \multicolumn{3}{l}{\textbf{Path Ranking}} \\ \cline{2-7}
                  &\textbf{MAE} &\textbf{MARE} &\multicolumn{1}{l|}{\textbf{MAPE}} & \textbf{MAE}    &\textbf{$\tau$} &\textbf{$\rho$} \\ \toprule[1pt]
\textbf{PIM-Temporal}  &42.27     &0.19    &\multicolumn{1}{l|}{27.95}       &0.19        &0.65           &0.70              \\ \hline
\textbf{WSCCL}   &\textbf{31.66}     &\textbf{0.13}      &\multicolumn{1}{l|}{\textbf{21.39}}      &\textbf{0.15}  &\textbf{0.68}     &\textbf{0.72}       \\ \toprule[2pt] 

\multirow{3}{*}{\textbf{Methods}} & \multicolumn{6}{l}{\textbf{Harbin}}                                                    \\ \cline{2-7} 
\multirow{2}{*}{} & \multicolumn{3}{l|}{\textbf{Travel Time Estimation}} & \multicolumn{3}{l}{\textbf{Path Ranking}} \\ \cline{2-7}
&\textbf{MAE} &\textbf{MARE} &\multicolumn{1}{l|}{\textbf{MAPE}} & \textbf{MAE}    &\textbf{$\tau$} &\textbf{$\rho$} \\ \toprule[1pt]
\textbf{PIM-Temporal} &190.25       &0.20   &\multicolumn{1}{l|}{21.29}          &0.20       &0.47                 &0.50              \\ \hline
\textbf{WSCCL}    &\textbf{178.89}     &\textbf{0.18}   &\multicolumn{1}{l|}{\textbf{19.43}}          &\textbf{0.14}      &\textbf{0.68}                &\textbf{0.73}                  \\ \toprule[2pt] 

\multirow{3}{*}{\textbf{Methods}} & \multicolumn{6}{l}{\textbf{ChengDu}}                                                    \\ \cline{2-7} 
\multirow{2}{*}{} & \multicolumn{3}{l|}{\textbf{Travel Time Estimation}} & \multicolumn{3}{l}{\textbf{Path Ranking}} \\ \cline{2-7}
&\textbf{MAE} &\textbf{MARE} &\multicolumn{1}{l|}{\textbf{MAPE}} & \textbf{MAE}    &\textbf{$\tau$} &\textbf{$\rho$} \\ \toprule[1pt]
\textbf{PIM-Temporal} &288.63       &0.30  &\multicolumn{1}{l|}{35.37}         &0.19       &0.79                &0.82              \\ \hline
\textbf{WSCCL}    &\textbf{281.20} &\textbf{0.29} &\multicolumn{1}{l|}{\textbf{33.30}}          &\textbf{0.13}      &\textbf{0.84}                &\textbf{0.86}                \\ \toprule[2pt] 
\end{tabular}

\label{tb:utpr}
\end{table}

\begin{table}[tp]
	\caption{{Comparison with Supervised Methods}}
	\centering
	\begin{tabular}{l|llllll}
		\toprule[2pt]
		\multirow{3}{*}{\textbf{Methods}} & \multicolumn{6}{l}{\textbf{Aalborg}}                                                    \\ \cline{2-7} 
		\multirow{2}{*}{} & \multicolumn{3}{l|}{\textbf{Travel Time Estimation}} & \multicolumn{3}{l}{\textbf{Path Ranking}} \\ \cline{2-7}
		&\textbf{MAE} &\textbf{MARE} &\multicolumn{1}{l|}{\textbf{MAPE}} & \textbf{MAE}    &\textbf{$\tau$} &\textbf{$\rho$} \\ \toprule[1pt]
		\textbf{PathRank-PR}  &37.09    &0.16    &\multicolumn{1}{l|}{23.89}   &0.24       &0.58 &0.62                     \\ 
		\textbf{PathRank-TTE}  &55.08     &0.24    &\multicolumn{1}{l|}{36.71}       &0.23        &0.64           &0.68          \\ \hline
		\textbf{HMTRL-PR}  &40.59    &0.18    &\multicolumn{1}{l|}{21.81}   &0.25       &0.60 &0.64                     \\ 
		\textbf{HMTRL-TTE}  &47.22     &0.21    &\multicolumn{1}{l|}{29.97}       &0.17        &0.65           &0.68          \\ \hline
		\textbf{DeepGTT-PR}  &44.78    &0.20    &\multicolumn{1}{l|}{26.53}   &0.31       &0.56 &0.57                     \\
		\textbf{DeepGTT-TTE}  &59.52     &0.26    &\multicolumn{1}{l|}{37.80}       &0.39        &0.12           &0.12          \\ 
		\hline
		\textbf{WSCCL}   &\textbf{31.66}     &\textbf{0.13}      &\multicolumn{1}{l|}{\textbf{21.39}}      &\textbf{0.15}  &\textbf{0.68}     &\textbf{0.72}       \\ \toprule[2pt]

		\multirow{3}{*}{\textbf{Methods}} & \multicolumn{6}{l}{\textbf{Harbin}}                                                    \\ \cline{2-7} 
		\multirow{2}{*}{} & \multicolumn{3}{l|}{\textbf{Travel Time Estimation}} & \multicolumn{3}{l}{\textbf{Path Ranking}} \\ \cline{2-7}
		&\textbf{MAE} &\textbf{MARE} &\multicolumn{1}{l|}{\textbf{MAPE}} & \textbf{MAE}    &\textbf{$\tau$} &\textbf{$\rho$} \\ \toprule[1pt]
		\textbf{PathRank-PR}  &190.08    &0.20    &\multicolumn{1}{l|}{20.12}      &0.21        &0.36           &0.39                 \\ 
		\textbf{PathRank-TTE}  &204.49     &0.21    &\multicolumn{1}{l|}{23.87}       &0.18        &0.55           &0.60            \\ \hline
		\textbf{HMTRL-PR}  &228.58    &0.24    &\multicolumn{1}{l|}{23.60}      &0.21        &0.36           &0.41                 \\ 
		\textbf{HMTRL-TTE}  &260.65     &0.27    &\multicolumn{1}{l|}{31.56}       &0.22        &0.51           &0.56            \\ \hline
		
		\textbf{DeepGTT-PR}  &214.95    &0.22    &\multicolumn{1}{l|}{22.76}       &0.22        &0.38          &0.42            \\ 
		
		\textbf{DeepGTT-TTE}  &243.12    &0.25    &\multicolumn{1}{l|}{29.22}      &0.29        &0.04           &0.04                 \\ 
		\hline
		\textbf{WSCCL}    &\textbf{178.89}     &\textbf{0.18}   &\multicolumn{1}{l|}{\textbf{19.43}}          &\textbf{0.14}      &\textbf{0.68}                &\textbf{0.73}                  \\ \toprule[2pt] 
		
		\multirow{3}{*}{\textbf{Methods}} & \multicolumn{6}{l}{\textbf{ChengDu}}                                                    \\ \cline{2-7} 
		\multirow{2}{*}{} & \multicolumn{3}{l|}{\textbf{Travel Time Estimation}} & \multicolumn{3}{l}{\textbf{Path Ranking}} \\ \cline{2-7}
		&\textbf{MAE} &\textbf{MARE} &\multicolumn{1}{l|}{\textbf{MAPE}} & \textbf{MAE}    &\textbf{$\tau$} &\textbf{$\rho$} \\ \toprule[1pt]
		\textbf{PathRank-PR}  &334.94     &0.32     &\multicolumn{1}{l|}{35.11}      &0.22        &0.61           &0.62                \\ 
		\textbf{PathRank-TTE}  &368.71      &0.39     &\multicolumn{1}{l|}{47.65}       &0.17     &0.79          &0.83           \\ \hline
		\textbf{HMTRL-PR}  &360.08     &0.38     &\multicolumn{1}{l|}{37.33}      &0.26        &0.26           &0.24                 \\ 
		\textbf{HMTRL-TTE}  &372.08     &0.39    &\multicolumn{1}{l|}{18.08}       &0.16         &0.77          &0.79              \\ \hline
		
		\textbf{DeepGTT-PR}  &305.08     &0.33        &\multicolumn{1}{l|}{35.47}    &0.24   &0.19         &0.20            \\ 
		
		\textbf{DeepGTT-TTE}  &368.76    &0.39    &\multicolumn{1}{l|}{47.77}     &0.23        &0.20          &0.22              \\ 
		\hline
		\textbf{WSCCL}    &\textbf{281.20} &\textbf{0.29} &\multicolumn{1}{l|}{\textbf{33.30}}          &\textbf{0.13}      &\textbf{0.84}                &\textbf{0.86}                \\ \toprule[2pt] 
	\end{tabular}
	\label{tb:verc}
\end{table}

\begin{table}[ht]
\caption{Effects of $\lambda$}
\centering
\begin{tabular}{l|llllll}
\toprule[2pt]
\multirow{3}{*}{\textbf{$\lambda$}} & \multicolumn{6}{l}{\textbf{Aalborg}}\\ \cline{2-7} 
                        & \multicolumn{3}{l|}{\textbf{Travel Time Estimation}}  & \multicolumn{3}{l}{\textbf{Path Ranking}} \\ \cline{2-7} 
                        &\textbf{MAE} &\textbf{MARE} &\multicolumn{1}{l|}{\textbf{MAPE}} & \textbf{MAE}    &$\tau$ &$\rho$        \\ \toprule[1pt]
0.0  &51.19    &0.22    &\multicolumn{1}{l|}{31.13}       &0.24       &0.54         &0.58                          \\ \hline
0.2  &40.25    &0.18    &\multicolumn{1}{l|}{24.67}       &0.22       &0.60         &0.64                          \\ \hline
0.4  &34.22    &0.15    &\multicolumn{1}{l|}{21.80}       &0.18       &0.64         &0.68                   \\ \hline
0.6  &34.76    &0.15    &\multicolumn{1}{l|}{22.35}       &0.17       &0.65         &0.69                         \\ \hline
0.8   &\textbf{31.66}    &\textbf{0.14}    &\multicolumn{1}{l|}{\textbf{21.39}}       &\textbf{0.15}       &\textbf{0.68}         &\textbf{0.72}                   \\ \hline 
1.0    &32.80    &0.14    &\multicolumn{1}{l|}{23.34}       &0.21       &0.57         &0.62 \\ \toprule[2pt]
\end{tabular}
\label{tb1:lam}
\end{table}

\begin{table}[!ht]
\caption{Effects of Number of Meta-Set. }
\centering
\begin{tabular}{l|llllll}
\toprule[2pt]
\multirow{3}{*}{$N$} & \multicolumn{6}{l}{\textbf{Aalborg}}                                                    \\ \cline{2-7} 
\multirow{2}{*}{} & \multicolumn{3}{l|}{\textbf{Travel Time Estimation}} & \multicolumn{3}{l}{\textbf{Path Ranking}} \\ \cline{2-7}
                  &\textbf{MAE} &\textbf{MARE} &\multicolumn{1}{l|}{\textbf{MAPE}} & \textbf{MAE}    &\textbf{$\tau$} &\textbf{$\rho$} \\ \toprule[1pt]
2       &36.64     &0.17    &\multicolumn{1}{l|}{25.86}       &0.20       &0.56            &0.60    \\ \hline
6  &36.06     &0.16    &\multicolumn{1}{l|}{24.96}       &0.20        &0.56           &0.60                            \\ \hline
10  &\textbf{31.66}     &\textbf{0.14}      &\multicolumn{1}{l|}{\textbf{21.39}}      &\textbf{0.15}  &\textbf{0.68}     &\textbf{0.72}                     \\ \hline
14  &33.16     &0.15      &\multicolumn{1}{l|}{21.60}      &0.19  &0.58     &0.63  \\ \hline
18  &33.47     &0.15     &\multicolumn{1}{l|}{21.65}      &0.20  &0.56     &0.61 \\ \toprule[2pt]

\multirow{3}{*}{\textbf{Methods}} & \multicolumn{6}{l}{\textbf{Harbin}}                                                    \\ \cline{2-7} 
\multirow{2}{*}{} & \multicolumn{3}{l|}{\textbf{Travel Time Estimation}} & \multicolumn{3}{l}{\textbf{Path Ranking}} \\ \cline{2-7}
&\textbf{MAE} &\textbf{MARE} &\multicolumn{1}{l|}{\textbf{MAPE}} & \textbf{MAE}    &\textbf{$\tau$} &\textbf{$\rho$} \\ \toprule[1pt]
2   &201.97    &0.21   &\multicolumn{1}{l|}{22.60}       &0.19       &0.54                 &0.58               \\ \hline
6   &210.78      &0.22   &\multicolumn{1}{l|}{23.36}          &0.19       &0.52                 &0.57                 \\ \hline
10   &\textbf{178.89}     &\textbf{0.18}   &\multicolumn{1}{l|}{\textbf{19.43}}          &\textbf{0.14}      &\textbf{0.68}                &\textbf{0.73}                 \\ \hline 
14   &199.92     &0.21   &\multicolumn{1}{l|}{22.02}         &0.17     & 0.58              &0.62               \\ \hline

18   &201.79     &0.21   &\multicolumn{1}{l|}{22.26}          &0.17     &0.56               &0.61               \\ \toprule[2pt]
\end{tabular}
\label{tb:num}
\end{table}

\subsubsection{\textbf{Comparison with Temporally Enhanced Unsupervised Method}}~To compare \emph{WSCCL} with the unsupervised \emph{PIM} method, we first incorporate a temporal representation into the non-temporal path representations learned by \emph{PIM}. 
Specifically, we use the same temporal embedding to learn temporal representations and then concatenate these with the path representation from PIM to obtain~\emph{PIM-Temporal} unsupervised TPRs. 
The results of comparing this approach with \emph{WSCCL} are reported in Table~\ref{tb:utpr}. We see that \emph{WSCCL} outperforms \emph{PIM-Temporal} on both tasks.
This indicates that TPRs obtained by adding a temporal representation to the path representation directly is not as good as the TPR learned by \emph{WSCCL}. This is because the added temporal representation can only capture the overall traffic condition on the road network for all the paths, yet not more unique spatio-temporal path representations learned by our \emph{Temporal Path Encoder} for different paths. 
This experiment shows that it is not feasible to obtain a generic TRP by independently adding a temporal representation to an unsupervised learned generic PR~(i.e., spatial representation). Further, it offers evidence of a more correlated and intricate interplay between space and time in TPRs. For example, during the morning peak hours, different paths may have different traffic conditions. 

\subsubsection{\textbf{Comparison with Supervised Method}} To study the applicability  of TPRs from supervised method across tasks, we use the supervised methods \emph{PathRank},~\emph{HMTRL} and \emph{DeepGTT} as baselines. In the supervised methods, we define a primary and a secondary task: A supervised model is trained on the primary task, and then the learned path representation is applied to the secondary task directly. Thus, we have two experimental settings: 1)~\emph{Baseline-PR}, travel-time estimation is the primary task and path ranking is the secondary task; 2)~\emph{Baseline-TTE}, where path ranking is the primary task and travel-time estimation is the secondary task. 
The results are reported in Table~\ref{tb:verc}. 
We first observe that \emph{WSCCL} achieves the best performance on both downstream tasks. 
Further, we observe that the performance of \emph{PathRank} and \emph{HMTLR} are always better on the primary task than on the secondary task. For example, for travel-time estimation on Aalborg, \emph{PathRank-PR} is better than \emph{PathRank-TTE}. This is evidence of the drawbacks of supervised approaches that task-specific TPRs do not generalize well across tasks. 
Moreover, we also observe that \emph{DeeGTT-PR} is always better than \emph{DeeGTT-TTE} on both data sets. This is because \emph{DeepGTT} is designed to do travel time distribution estimation. Given the task of path ranking, whose distribution may not follow the same inverse-Gaussian distribution like in travel time, so it fails in this case.

\subsubsection{\textbf{Parameter Studies}}
We study the effects of $\lambda$ and $N$. 
\noindent
\paragraph{Effects of $\lambda$}
To study the effect of the balancing factor $\lambda $~(cf. Eq.~\ref{eq:jl}), we conduct a parameter study on Aalborg. Based on the results reported in Table~\ref{tb1:lam}, we see that the performance of our model changes when varying $\lambda$. We can also see that the optimal $\lambda$ is 0.8, which means that both global WSC loss and local WSC loss can contribute to the model's performance. 
When $\lambda =0.0$, the global WSC loss is ignored, which yields poor performance. When $\lambda =1.0$, the local contrastive loss is ignored, and the best performance is not obtained, although the performance is quite good. 
When $\lambda > 0 $, meaning that we consider both weakly-supervised and local WSC loss, we observe that the prediction performance is improved. 
Overall, we conclude that global WSC loss is more important than the local loss. 

\noindent
\paragraph{Effects of $N$}
To study the effect of varying the number of \textit{Experts} $N$, which is also the number of curriculum stages as we always set $N=M$, in the curriculum strategy, we observe the performance for values of $N$ in $\{2,6,10,14,18\}$. The results in Table~\ref{tb:num} indicate that the best performance is obtained for $N=10$ on Aalborg and Harbin data sets. We also observe that when $N$ is too small, the curriculum strategy is not effective. 
This occurs because the number of \textit{Experts} is small, meaning that the difficulty scores have more uncertainty and inaccuracy. This can also be the reason why the number of curriculum stages is small in the curriculum sample selection stage, such that the samples in the beginning are difficult to learn. Next, when the $N$ becomes too large, this may cause problems in the curriculum sample selection stage because the diversity and number of data points in the meta-sets may be too small, resulting in the model overfitting at each stage. 

%
\section{Conclusions and Future Work}
\label{sec:con}
We study temporal path representation learning using weak labels. We propose a novel weakly supervised contrastive learning method that uses weakly supervised contrastive learning and local constrative loss. Next, we integrate curriculum learning into the method to further enhance its performance. Finally, we report on experiments on three data sets in the settings of three downstream tasks, finding that our proposal achieves significant performance improvements over unsupervised and supervised baselines. In addition, the proposed method can be utilized as a pre-training method to enhance supervised temporal path representation learning. As future work, it is of interest to study how to incorporate additional weak labels such as drivers and vehicle types. 

\section*{Acknowledgments}

This work was partially supported by Independent Research Fund Denmark under agreements 8022-00246B and 8048-00038B, the VILLUM FONDEN under agreements 34328 and 40567, and the Innovation Fund Denmark centre, DIREC. 

\balance
\bibliographystyle{IEEEtran}
\bibliography{IEEEexample}

\end{document}